\documentclass[3p,a4paper]{elsarticle}
\usepackage[utf8]{inputenc}
\usepackage{subfigure}
\usepackage[ruled,vlined,linesnumbered]{algorithm2e}
\usepackage{amsmath,amsfonts}
\usepackage{amsthm}
\title{Symbolic--KAN: Kolmogorov-Arnold Networks with Discrete Symbolic Structure for Interpretable  Learning}

\author[myUaddress]{Salah A Faroughi \corref{mycorrespondingauthor}}

\author[myUaddress]{Farinaz Mostajeran}

\author[myUaddress3b,myUaddress3,myUaddress3c]{Amirhossein Arzani}

\author[myUaddress2]{Shirko Faroughi}

\cortext[mycorrespondingauthor]{Corresponding author: salah.faroughi@utah.edu}
\address[myUaddress]{Energy \& Intelligence Lab, Department of Chemical Engineering, University of Utah, Salt Lake City, Utah,  84112, USA
}
\address[myUaddress3b]{Department of Mechanical Engineering, University of Utah, Salt Lake City, Utah,  84112, USA
}

\address[myUaddress3]{Scientific Computing and Imaging Institute, University of Utah, Salt Lake City, Utah,  84112, USA
}

\address[myUaddress3c]{Department of Biomedical Engineering, University of Utah, Salt Lake City, Utah,  84112, USA
}

\address[myUaddress2]{Department of Mechanical Engineering, School of Engineering, Urmia University of Technology, Urmia, Iran
}

\date{\today}

\usepackage{setspace} 
\usepackage{tabularx}
\usepackage{lineno,hyperref}
\usepackage{amsmath}
\usepackage{bm}
\usepackage{amssymb}
\usepackage[]{amsmath}
\usepackage{amsthm}
\usepackage{upgreek}
\usepackage{float}

\usepackage{pgfplots}
\pgfplotsset{compat=1.5}
\usepgfplotslibrary{units}
\usetikzlibrary{spy}
\usetikzlibrary{pgfplots.groupplots}

\usepackage{epstopdf}
\usepackage{units}
\usepackage{lscape}
\usepackage{booktabs}
\usepackage{gensymb}
\usepackage{multirow}

\usepackage{graphicx}
\usepackage{caption}
\usepackage{subcaption}

\usepackage{booktabs}
\usepackage{multirow}
\usepackage{caption}
\usepackage{subcaption}
\biboptions{numbers,sort,compress} 
\usepackage{soul,color}
\soulregister\cite7
\soulregister\ref7
\soulregister\pageref7

\usepackage[flushleft]{threeparttable}
\usepackage{tikz}
\usepackage{placeins}
\usepackage{color}
\usepackage{hyperref}

\usepackage{array}
\usepackage{tabularx}
\usepackage{pdfpages}
\usepackage{natbib}
\usepackage{textcomp, gensymb}
\usepackage{booktabs} 

\newtheorem{open problem}[theorem]{Open Problem}

\begin{document}

\begin{abstract}

Symbolic discovery of governing equations is a long-standing goal in scientific machine learning, yet a fundamental trade-off persists between interpretability and scalable learning. Classical symbolic regression methods yield explicit analytic expressions but rely on combinatorial search, whereas neural networks scale efficiently with data and dimensionality but produce opaque representations. In this work, we introduce Symbolic Kolmogorov-Arnold Networks (Symbolic-KANs), a neural architecture that bridges this gap by embedding discrete symbolic structure directly within a trainable deep network. Symbolic-KANs represent multivariate functions as compositions of learned univariate primitives applied to learned scalar projections, guided by a library of analytic primitives, hierarchical gating, and symbolic regularization that progressively sharpens continuous mixtures into one-hot selections. After gated training and discretization, each active unit selects a single primitive and projection direction, yielding compact closed-form expressions without post-hoc symbolic fitting. Symbolic-KANs further act as scalable primitive discovery mechanisms, identifying the most relevant analytic components that can subsequently inform candidate libraries for sparse equation-learning methods. We demonstrate that Symbolic-KAN reliably recovers correct primitive terms and governing structures in data-driven regression and inverse dynamical systems. Moreover, the framework extends to forward and inverse physics-informed learning of partial differential equations, producing accurate solutions directly from governing constraints while constructing compact symbolic representations whose selected primitives reflect the true analytical structure of the underlying equations. These results position Symbolic-KAN as a step toward scalable, interpretable, and mechanistically grounded learning of governing laws.

\end{abstract}


\maketitle

\section{Introduction}\label{sec:Intro}
Scientific machine learning (SciML) has become a central paradigm for building predictive and surrogate models in science and engineering, where the goal is often to learn complex constitutive relations, closure laws, or operators directly from data while respecting physical structure and constraints~\cite{cuomo2022scientific, MAO2020112789, Li2020FNO,faroughi2024SciMLRev}. Deep neural networks, and in particular multilayer perceptrons (MLPs), have enabled impressive advances in this direction, but their success has also amplified concerns about opacity, lack of trust, and difficulty in diagnosing or correcting failure modes \cite{faroughi2024SciMLRev}. In many applications, domain experts need not only accurate predictions but also compact, human-readable expressions that can be inspected, analyzed, and embedded into existing (mechanistic) modeling pipelines ~\cite{Ljung2010,cranmer2023interpretable,ADAMSINDy}. This motivates symbolic regression approaches as well as neural architectures that can admit symbolic interpretation, rather than relying solely on post–hoc explanations \cite{camburu2020explaining,samek2021,bordt2022post}. Kolmogorov-Arnold networks (KANs)  offer a promising step in this direction \cite{faroughi2026KanRev,cruz2025state}. By construction,  they parameterize a multivariate mapping as a superposition of trainable univariate functions and linear combinations, echoing the Kolmogorov-Arnold representation theorem~\cite{Arnold1958, ismailov2014approximation, ismailov2023three, schmidt2021kolmogorov, poluektov2023construction}. Because KANs learn explicit one-dimensional basis function, often implemented as splines or other simple function families and polynomials, their internal structure is much closer to an analytic formula than that of conventional MLPs~\cite{liu2024kan, liu2024kan2, yu2024kan, zeng2024kan, faroughi2025neural}, suggesting that KANs could naturally bridge neural modeling and symbolic regression.

To address the needs for symbolic output to uncover governing laws directly from observations,  a rich body of interpretable, data-driven modeling approaches has emerged beyond traditional system identification pipelines \cite{Ljung2010}. Symbolic regression methods, particularly those based on genetic algorithms, search over compositions of mathematical operators to construct compact analytic expressions of dynamical systems, enabling flexible model discovery without requiring prior parametric assumptions~\cite{Quade2016, Gondhalekar2009}. Although powerful, these strategies often suffer from high computational cost and phenomena such as expression bloat, which can reduce interpretability and generalization~\cite{Quade2016,schmidt2020nonparametric,cranmer2023interpretable}. Complementing symbolic regression, sparsity-promoting library-based frameworks such as the Sparse Identification of Nonlinear Dynamics (SINDy) leverage libraries of candidate functions and sparse regularization to identify parsimonious governing equations~\cite{Brunton2016}. 
SINDy has been applied successfully in diverse settings, e.g.,  fluid dynamics, chaotic dynamical systems, and chemical processes~\cite{fukami2021sparse,kaptanoglu2023benchmarking,yang2020rapid,hoffmann2019reactive}. 
It has also been extended for active learning and control~\cite{fasel2022ensemble}, stochastic dynamics~\cite{boninsegna2018sparse}, nonlinear reduced-order modeling~\cite{csala2026decomposed}, equation-based operator learning~\cite{arzani2025interpreting}, and Bayesian learning~\cite{champneys2025bindy} (among other developments~\cite{brunton2025machine}).  These advances offer interpretable and computationally efficient alternatives to symbolic regression when suitable candidate terms are known. Recent enhancements, such as ADAM-SINDy~\cite{ADAMSINDy}, integrate SINDy with modern gradient-based optimization to simultaneously infer coefficients and nonlinear parameters, e.g., frequencies, exponential rates, and non-integer exponents, addressing key limitations of SINDY's classical library-based search, which require prior knowledge of nonlinear parameters. A major limitation of these methods is that they cannot inherently generate new combinations of candidate terms from the library, which means important structures are overlooked unless they are explicitly included as individual terms in that library.  Recent advances attempt to alleviate these issues from complementary directions. SINDy-KANs~\cite{howardsindy} embed sparse regression within Kolmogorov--Arnold networks, enabling hierarchical discovery of nonlinear compositions while enforcing sparsity at the level of individual activation functions, thereby reducing reliance on large global candidate libraries. In parallel, State-Space KANs~\cite{cruz2025state} integrate KANs into structured dynamical models, modeling nonlinear state transitions within a physically meaningful state-space framework and improving interpretability through decomposed univariate representations, while operating directly in discrete time without requiring numerical differentiation. Such combinations point toward more parsimonious, interpretable, and mechanistically grounded representations, with improved alignment to the underlying physics, particularly when the governing structure is not known \emph{a priori}. 

KANs  have recently emerged as a flexible alternative to MLPs in several major SciML workflows, including purely data--driven learning, physics--informed training, and neural operator learning \cite{liu2024kan, liu2024kan2, koenig2024kan, yu2024kan, zeng2024kan, wang2025efkan, jacob2024spikans, faroughi2025neural}. Our recent review on KANs versus MLPs \cite{faroughi2026KanRev} clearly highlights both the strengths and the limitations of KANs. In the data--driven setting, KAN variants have been shown to achieve improved accuracy, data efficiency, and robustness across regression, classification, and sequence--modeling benchmarks by replacing fixed nonlinear activations with learned univariate bases along each edge~\cite{chen2024sckansformer, li2024hkan, zhou2025askan, ta2025prkan, mostajeran2026epi}. Physics--informed KANs (PIKANs) embed governing equations, residuals, or conservation laws directly into the training objective, leveraging the structured representation to improve stability and convergence relative to PINNs based on MLPs~\cite{xiapikans, mostajeran2025scaled, daryakenari2025representation, cheon2024kolmogorov,toscano2025kkans}. In parallel, operator--learning architectures that integrate KAN blocks into DeepONet--like or Fourier--operator frameworks have demonstrated competitive or superior performance to classical neural operators, particularly on problems with strong multiscale structure or irregular inputs~\cite{abueidda2025deepokan, xu2024fourierkan}. Across these regimes, comparisons against MLP--based PINNs, DeepONets, and related baselines indicate that KANs can reduce training stiffness, mitigate spectral bias, and provide smoother function approximations~\cite{faroughi2026KanRev,faroughi2025neural,liu2024kan}.  The expressive power of KANs is driven by their choice of univariate basis functions and associated parameterizations \cite{faroughi2026KanRev}. Existing KAN architectures predominantly rely on B-spline or piecewise-polynomial primitives, often implemented as trainable spline bases with learnable knot locations or adaptive weights~\cite{liu2024kan, liu2024kan2, koenig2024kan, li2024hkan, wang2025efkan}. Other variants incorporate Fourier-type primitives to better capture oscillatory structure~\cite{xu2024fourierkan}, or draw on wavelet constructions and multi-resolution analysis to represent localized, multiscale features~\cite{seydi2024unveiling,liang2026wkpnet}. While these design choices improve approximation quality and training dynamics, the resulting global expressions obtained by composing many learned basis functions can quickly become long and unwieldy. Even though each primitive is individually simple, the overall functional form represented by a deep KAN can be as opaque as that of a standard MLP when written out symbolically, which limits its usefulness as a genuinely interpretable or mechanistic model. In practice, directly reading off a compact governing equation or constitutive relation from a trained basic KAN (i.e., using a fixed form of learnable basis information) remains challenging, especially in higher-dimensional settings.

Motivated by the aforementioned challenges in sparsity-based methods and baseline KANs, we introduce a symbolic Kolmogorov--Arnold Network (Symbolic--KAN) framework that bridges these two paradigms. Symbolic--KAN is a neural architecture that embeds discrete symbolic structure directly within a trainable deep network. It models multivariate functions as compositions of learned univariate primitives acting on learned scalar projections, guided by a library of analytic primitives, a hierarchical gating mechanism, and symbolic regularization that progressively refines soft mixtures into one-hot selections (i.e., low-complexity representations). After gated training and subsequent discretization, each active unit commits to a single primitive and a single projection direction, yielding compact closed-form expressions without requiring post-hoc symbolic regression. Symbolic--KAN replaces conventional KAN formulations that rely on fixed basis functions~\cite{faroughi2026KanRev}, such as B-splines~\cite{yu2024kan, ta2024bsrbf} or Chebyshev polynomials~\cite{mostajeran2025scaled}, with a formulation in which candidate primitives are dynamically refined and sparsity is enforced over a reduced, data-informed functional space. This results in discrete, low-complexity symbolic surrogates aligned with the Kolmogorov--Arnold representation. In this way, Symbolic--KAN acts as a scalable mechanism for discovering informative primitives, isolating key analytic components that can subsequently be assembled into candidate libraries for sparse equation-learning frameworks when appropriate structures are hypothesized (i.e., serving as a pre-library selector for sparsity-based methods). We demonstrate that Symbolic--KAN recovers correct primitive terms in data-driven regression and identifies governing structures in dynamical systems. Furthermore, the framework extends to physics-informed learning of partial differential equations in both forward and inverse settings, where it learns accurate solutions directly from data and governing constraints while producing compact symbolic surrogates that reflect the analytical structure of the solution and/or underlying PDEs. While Symbolic--KAN improves structural identifiability and yields clearer functional forms, it does not fully resolve the long-standing challenges of neural network extrapolation and out-of-distribution generalization~\cite{Wang2020NTKPinns, JAGTAP2022111402, cranmer2023interpretable, arzani2025interpreting}. Rather, it provides a principled step toward models that are both scalable and amenable to symbolic inspection, laying the groundwork for future developments in robust, interpretable neural solvers and operator learning. The formulation of Symbolic--KAN is detailed in Section~\ref{sec:method}, results are presented in Section~\ref{sec:results}, and concluding remarks are provided in Section~\ref{sec:conclusion}.

\section{Symbolic Kolmogorov--Arnold Networks}\label{sec:method}

In this section, we introduce the proposed Symbolic Kolmogorov--Arnold network (Symbolic--KAN) framework and describe its architectural and training components. The formulation begins by defining the overall network representation and its relation to the Kolmogorov--Arnold representation theory. We then detail the construction of individual units, which combine learnable scalar projections with a library of analytic univariate primitives. To enable interpretable structure discovery, we present a gated training strategy that gradually transforms soft combinations of primitives into discrete selections. Finally, we present the loss function and the complete training pipeline employed to learn compact symbolic representations of the underlying data or governing equations.

\subsection{Network Formulation}

Symbolic--KAN follows the Kolmogorov--Arnold representation theorem (KART) \cite{lai2021kolmogorov,fakhoury2022exsplinet,he2023optimal} that states any continuous multivariate function 
$F(\boldsymbol{\xi}) : [0,1]^n \rightarrow \mathbb{R}$ can be represented as a finite superposition of univariate functions and additions. 
In one canonical form, one writes,
\begin{equation}
F(\boldsymbol{\xi} = (x_1,\dots,x_n))
=
\sum_{i=1}^{2n+1}
\Phi_i\!\left(
\sum_{j=1}^{n}
\psi_{ij}(x_j)
\right),
\label{eq:kart}
\end{equation}
where $\psi_{ij} : [0,1] \rightarrow \mathbb{R}$ and $\Phi_i : \mathbb{R} \rightarrow \mathbb{R}$ are the  inner and outer continuous univariate functions, respectively, and $\boldsymbol{\xi} \in \mathbb{R}^n$ indicates  the input vector.  Equation~\eqref{eq:kart} shows that the high-dimensional mapping $F$ can be decomposed into: (i) inner sums that compress the $n$-dimensional input into a set of scalar latent variables, and (ii) outer nonlinearities $\Phi_i$ that act on those scalar combinations, followed by a final summation across the outer index $i$.  Importantly, KART is an existence theorem and so it guarantees that such a representation is possible, but it does not prescribe unique or constructive forms for $\Phi_i$ and $\psi_{ij}$. A multilayer Symbolic--KAN takes Eq.~\eqref{eq:kart} as an architectural blueprint and rewrite it as,
\begin{equation}  
F_{_\text{KAN}}(\boldsymbol{\xi}) = (\boldsymbol{\Phi}_L \circ \boldsymbol{\Phi}_{L-1} \circ \cdots \circ \boldsymbol{\Phi}_2 \circ ~ \boldsymbol{\Phi}_1 )(\boldsymbol{\xi}), 
\end{equation}
where each layer function, denoted as $\boldsymbol{\Phi}_\ell$, is made up of learnable univariate functions that link to the subsequent layer ${\ell+1}$, and $\circ$ represents the composition of functions. In a standard KAN \cite{faroughi2026KanRev}, for a layer $\ell$ with $n_{\ell}$ output nodes feeding into the next layer, which contains 
$n_{\ell+1}$ nodes, the transformation between these two layers can be written as a matrix of 
learnable univariate functions,
\begin{equation}
\boldsymbol{\Phi}^{\text{(Standard~KAN)}}_{\ell} =
\begin{pmatrix} 
\psi^{(\ell)}_{1,1}(\cdot) & \psi^{(\ell)}_{1,2}(\cdot) & \cdots & \psi^{(\ell)}_{1,n_{\ell}}(\cdot) \\
\psi^{(\ell)}_{2,1}(\cdot) & \psi^{(\ell)}_{2,2}(\cdot) & \cdots & \psi^{(\ell)}_{2,n_{\ell}}(\cdot) \\
\vdots & \vdots & \ddots & \vdots \\
\psi^{(\ell)}_{n_{\ell+1},1}(\cdot) & \psi^{(\ell)}_{n_{\ell+1},2}(\cdot) & \cdots & \psi^{(\ell)}_{n_{\ell+1},n_{\ell}}(\cdot)
\end{pmatrix} \;,
\label{PHI_matrx} 
\end{equation}
in which each univariate function denoted as $\psi^{(\ell)}_{i,j}(\cdot)$ is parameterized by a chosen basis function \cite{faroughi2026KanRev}. Equation \eqref{PHI_matrx}  shows that the standard KAN adheres closely to the canonical KART construction by assigning a distinct univariate map $\psi^{(\ell)}_{i,j}$ to every coordinate $j$ feeding into an output node $i$ at layer $\ell$. 

\begin{figure}[!ht]
    \centering
    \includegraphics[width=1.0\linewidth]{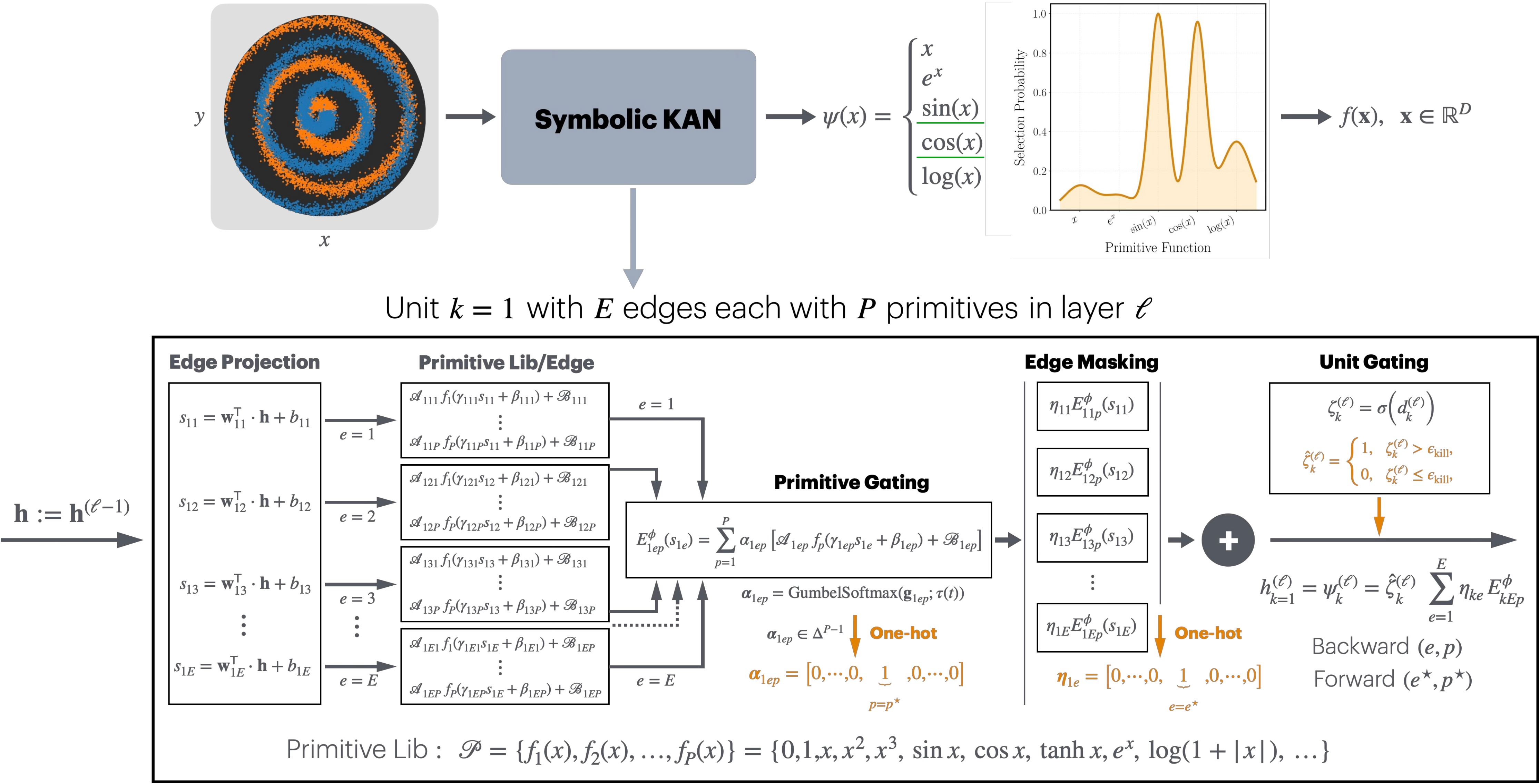}
    \caption{A schematic layout of the operations performed by a unit within a layer in the proposed Symbolic--KAN framework.}
    \label{fig:sumKAN_ARC}
\end{figure}

However, in Symbolic--KAN's architecture we propose a change as shown in  Fig.~\ref{fig:sumKAN_ARC}. Symbolic--KAN preserves the KART principle that multivariate mappings arise from compositions of univariate functions, but adopts a more flexible and learnable inner decomposition. Instead of storing a full row of functions $\{\psi^{(\ell)}_{i,1},\ldots,\psi^{(\ell)}_{i,n_\ell}\}$, each unit $k$ at layer $\ell$ learns a scalar coordinate through projection, $s^{(\ell)}_k$, which replaces the fixed KART inner sum. A single univariate map,
$\
\psi^{(\ell)}_{k}(s^{(\ell)}_k),
$
is then applied to this learned one-dimensional projection. This collapses the two-index structure $(i,j)$ into a single symbolic index $k$, while still ensuring that each unit computes a \emph{univariate} transformation of a \emph{scalar} argument, which is the defining requirement of Kolmogorov-Arnold representations. Under this formulation, the Symbolic--KAN layer operator becomes,
\begin{equation}
\boldsymbol{\Phi}^{\text{(Symbolic--KAN)}}_\ell =
\left[ \psi^{(\ell)}_{1}(s^{(\ell)}_1), \,
        \cdots ,\, 
         \psi^{(\ell)}_{k}(s^{(\ell)}_k), \,
         \cdots ,\,
         \psi^{(\ell)}_{K_\ell}(s^{(\ell)}_{K_\ell})
         \right]^{\text{T}}, \qquad \quad  k \in \{1,\dots,K_\ell\},
\label{PHI_matrx_symKAN}
\end{equation}
where $K_l$ is total number of units in a layer, and the superscript \(\mathrm{T}\) denotes the transpose operator, and \(\boldsymbol{\Phi}^{\text{(Symbolic--KAN)}}_\ell\)
is a row vector whose entries correspond to the outputs of single univariate functions applied to learned scalar projections, rather than collections of coordinatewise univariate maps.
This reparametrization remains consistent within the Kolmogorov--Arnold principle \cite{kolmogorov1957representations,kolmogorov1961representation,li2024kolmogorov}, but grants the network the flexibility to discover meaningful combinations of upstream features and to represent them through interpretable symbolic primitives.

\subsection{Unit Construction}

Consider a fixed layer $\ell$ and a unit index $k \in \{1,\dots,K_\ell\}$.  This unit receives the previous layer activations $\boldsymbol{h}^{(\ell-1)} \in \mathbb{R}^{H_{\ell-1}}$ and is equipped with $E$ scalar ``edges'' indexed by $e \in \{1,\dots,E\}$.  Each edge in that unit first computes a scalar projection,
\begin{equation}
s_{ke}
=
\boldsymbol{w}_{ke}^{\mathsf{T}} \cdot \boldsymbol{h}^{(\ell-1)} + b_{ke},
\qquad
\boldsymbol{w}_{ke} \in \mathbb{R}^{H_{\ell-1}},\quad
b_{ke} \in \mathbb{R},
\label{eq:s_km}
\end{equation}
where the vectors $\boldsymbol{w}_{ke} \in \mathbb{R}^{H_{\ell-1}}$ and scalars $b_{ke}$ play the role of the inner linear combinations. This projection can be interpreted as a learned one-dimensional ``coordinate'' in the space of upstream features, and makes each edge's input a unique projection of inputs. Next, each edge transformation within unit $k$, i.e., $E^\phi$, is calculated as,
\begin{equation}\label{edge_transformation}
E^\phi_{keP}(s_{ke}) 
= 
\sum_{p=1}^{P} \alpha_{kep}\,
\big[
\mathcal{A}_{kep}\,f_p(\gamma_{kep}s_{ke} + \beta_{kep})
+ \mathcal{B}_{kep}
\big],
\end{equation}
where $P$ is the number of primitives in a diverse, interpretable library of analytic primitives, expressed as,
\begin{equation}
\mathcal{P}
=
\{ f_1(x), f_2(x), \dots, f_P(x) \}
=
\{0,1,x,\, x^2,\, x^3,\, \sin x,\, \cos x,\, \tanh x,\, e^x,\, \log(1+|x|),\, \ldots \}.
\end{equation}

In Eq.~\eqref{edge_transformation}, $\gamma_{kep}$ and $\beta_{kep}$ are scalar affine-in and $\mathcal{A}_{kep}$ and  $\mathcal{B}_{kep}$ are scalar affine-out parameters for primitive $p$ on edge $(k,e)$ that can be labeled as trainable parameters. Note that the affine transformation applied inside each primitive, $\gamma\,s + \beta$, is in principle capable of absorbing the external projection $s = \boldsymbol{w}^{\mathsf{T}} \cdot \boldsymbol{h}^{(\ell-1)} + b$. However, we deliberately retain both parameterizations in the architecture because they serve distinct functional and optimization roles. The projection weights $(\boldsymbol{w},b)$ define how a unit aggregates information from all upstream features and therefore control the geometry of the input subspace on which each univariate primitive operates. In contrast, the internal parameters $(\gamma,\beta)$ modulate the local scaling and translation of the primitive itself, providing fine-grained control over its operating region without altering how upstream features are combined. Keeping both affine mappings therefore decouples feature selection from primitive shaping, which can improve optimization conditioning, allows consistent primitive parameterization across edges, and yields more stable and interpretable symbolic forms. Empirically, this separation enhances trainability, prevents parameter entanglement, and preserves the modular structure required for symbolic export, even though the two affine layers are mathematically redundant when viewed solely from a function-approximation perspective. 
Also, in Eq.~\eqref{edge_transformation}, we use $\alpha_{kep}$ to linearly combines primitives for each edge.  In the fully symbolic regime, the trainable coefficients $\alpha_{kep}$ are encouraged (and eventually forced) to be one-hot over $p$ so that each edge selects a single primitive (i.e., $p^\star$ plus its affine parameters) from the library $\mathcal{P}$.  

For unit $k$ at layer $\ell$, the $E$ edges produce scalar outputs,
\begin{equation}
y_{ke}
=
E^\phi_{kep}(s_{ke}) ,
\qquad e = 1,\dots,E,
\end{equation}
where $s_{ke}$ is given by Eq.~\eqref{eq:s_km} and $E^\phi_{kep}$ by Eq.~\eqref{edge_transformation}.  The unit then aggregates these edge-wise candidates via an edge-selection mask $\boldsymbol{\eta}_k = (\eta_{k1},\dots,\eta_{kE})$, where $\eta_{ke} \in [0,1]$ and,
\begin{equation}
\sum_{e=1}^{E} \eta_{ke} = 1,
\end{equation}
and hence the unit output is  computed as,
\begin{equation}
  h^{(\ell)}_k
=
\sum_{e=1}^{E} \eta_{ke}\, y_{ke}
=
\sum_{e=1}^{E} \eta_{ke}
\left\{
\sum_{p=1}^{P} \alpha_{kep}\,
\big[
\mathcal{A}_{kep}\,f_p(\gamma_{kep}s_{ke} + \beta_{kep})
+ \mathcal{B}_{kep}
\big]
\right\}.
\label{eq:node_output_full}
\end{equation}

Equation~\eqref{eq:node_output_full} makes explicit that each unit computes, in principle, a sparse sum of analytic primitives applied to learned scalar projections of the previous layer.  After training and hardening, both the primitive-selection coefficients $\alpha_{kep}$ and edge-selection mask $\boldsymbol{\eta}_k$ become (approximately) one-hot (i.e.\ there exists a unique $e^\star$ such that $\eta_{ke^\star} = 1$ and $\eta_{ke} = 0$ for all $e \neq e^\star$), so that $h^{(\ell)}_k$ is implemented by a \emph{single} primitive $f_{p^\star}$ applied to a \emph{single} projection $s_{ke^\star}$, modulated by its affine parameters. In the strict symbolic limit,  the unit output then reads as,
\begin{equation}
h^{(\ell)}_k
 = \psi^{(\ell)}_k =
\mathcal{A}_{ke^\star p^\star}\,f_{p^\star}(\gamma_{ke^\star p^\star}s_{ke^\star} + \beta_{kep^\star})
+ \mathcal{B}_{ke^\star p^\star}.
\label{eq:node_output_oen_hat}
\end{equation}

For scalar-valued outputs, the network output is obtained via a fixed linear readout of the final layer, which in this work is taken to be a simple summation of the last-layer activations to comply with the Kolmogorov–Arnold representation theory.

\subsection{Gated Training and Discrete Selections}

Symbolic--KAN employs a hierarchical gating mechanism to convert a continuously parameterized mixture model into a strictly symbolic architecture. The model contains three classes of gates that collectively determine the discrete functional structure learned during training: (i) primitive-selection gates that choose one analytic primitive per edge, (ii) edge-selection masks that choose one scalar projection (i.e., one edge) per unit, and (iii) unit gates that decide whether a hidden unit remains active or is pruned from the architecture. All three mechanisms are first implemented through continuous, differentiable relaxations, since their discrete counterparts are non-differentiable, and training gradually sharpens these relaxations toward discrete limits. At convergence, the primitive gates and edge-selection masks become one-hot, inactive units are suppressed, and the network reduces to a compact symbolic representation with exactly one primitive and one projection per surviving unit.

Primitive-selection gates are defined for each unit $k$ and each incident edge $e\in\{1,\dots,E\}$ by a vector of primitive logits, $\boldsymbol{g}_{kep}$, corresponding to the $P$ analytic primitives $\{f_1,\dots,f_P\}$. A continuous relaxation is obtained via the Gumbel--Softmax operation \cite{strypsteen2021end},
\begin{equation}
    \boldsymbol{\alpha}_{kep}
    =
    \operatorname{GumbelSoftmax}(\boldsymbol{g}_{kep};\tau(t)),
    \qquad 
    \boldsymbol{\alpha}_{kep} \in \Delta^{P-1},
\end{equation}
where $\tau>0$ is a temperature parameter controlling the smoothness of the resulting distribution. At high temperature, $\boldsymbol{\alpha}_{ke}$ represents a diffuse convex combination of primitives; as $\tau\to 0$, the distribution becomes sharply peaked. Entropy regularization is also used to further encourage each $\boldsymbol{\alpha}_{ke}$ to converge toward a one-hot vector as $\tau$ decreases, see Section \ref{loss_sec}.  Note that in $\tau(t)$, the variable $t$ is the training iteration. Initially, a large value $\tau(0)=\tau_{\mathrm{start}}$ encourages wide exploration of the primitive library. Over training, $\tau(t)$ is annealed to a small value $\tau_{\mathrm{end}}$, causing the distributions $\boldsymbol{\alpha}_{ke}$ to become sharply concentrated.  After sufficient sharpening has occurred (low $\tau$ and small entropy), a hardening step is performed. Each primitive gate is replaced with the one-hot vector that selects the maximal logit,
\begin{equation}
\boldsymbol{\alpha}_{kep}\leftarrow\mathrm{onehot}\!\bigl(\arg\max_p \boldsymbol{g}_{kep}\bigr),
\end{equation}
to be leveraged during forward path. 
Note that during backpropagation, gradients propagate through the continuous relaxation, while in the forward computation one may employ either the soft or hard variant of the Gumbel--Softmax.

Edge-selection masks are also defined for each unit in Symbolic--KAN, as shown in Fig.~\ref{fig:sumKAN_ARC}, that receives $E$ candidate scalar projections. For unit $k$, the model computes a confidence score $c_{ke}$ for each edge $e$,  taken as the maximum component of $\boldsymbol{\alpha}_{ke}$, i.e.\ $c_{ke}=\max_p\alpha_{kep}$. These scores encourage edge selection based on the sharpness of the associated primitive gate. The vector $\boldsymbol{c}_k=(c_{k1},\dots,c_{kE})$ is normalized by a softmax to obtain continuous edge-scores, $ \boldsymbol{S}_k$,
\begin{equation}
    \boldsymbol{S}_k = \operatorname{softmax}(\boldsymbol{c}_k),
\end{equation}
after which a straight-through top-$1$ operator produces a deterministic one-hot mask,
\begin{equation}
    \hat{\boldsymbol{\eta}}_k = \operatorname{Top1}(\boldsymbol{S}_k),
    \qquad 
    \hat{\boldsymbol{\eta}}_{k} \in \{0,1\}^E,
    \qquad 
    \sum_{e=1}^{E} \hat{\eta}_{ke} = 1,
\end{equation}
where the resulting vector $\hat{\boldsymbol{\eta}}_k$ is the discrete edge-selection mask for unit $k$, with components $\hat{\eta}_{ke}\in\{0,1\}$ satisfying $\sum_{e=1}^{E}\hat{\eta}_{ke}=1$, ensuring that exactly one incoming edge is selected for that unit.
In edge selection, we also use a  non-maximum suppression (NMS) penalty, as described in Section \ref{loss_sec}, to discourage different edges of the same unit from selecting identical primitives. The forward pass thus selects exactly one edge for each unit, while the backward pass uses the gradients of $\boldsymbol{S}_k$. Note that no temperature is applied to edge selection, because $\boldsymbol{S}_k$ depends on the sharpening of primitive gates, and so the edge mask becomes increasingly stable as the primitive gates approach one-hot vectors.

Unit-level gating form the final (optional) step in Symbolic--Kan that provides the mechanism for further structural sparsification. Each unit in layer $\ell$ is assigned a continuous activation gate,
\begin{equation}
    \zeta^{(\ell)}_{k} = \sigma\!\left(d^{(\ell)}_{k}\right),
    \qquad
    \zeta^{(\ell)}_{k} \in (0,1),
\end{equation}
where $d^{(\ell)}_{k}$ is a trainable scalar logit.  
The unit’s pre-activation is modulated multiplicatively,
\begin{equation}
    \tilde{h}^{(\ell)}_{k}
    =
    \zeta^{(\ell)}_{k}\, h^{(\ell)}_{k},
\end{equation}
so that units with $\zeta^{(\ell)}_{k}$ close to zero contribute negligibly to the computation. During symbolic inference, the continuous gates are converted into discrete indicators via a hard thresholding rule,
\begin{equation}
    \hat{\zeta}^{(\ell)}_{k} =
    \begin{cases}
        1, & \zeta^{(\ell)}_{k} > \epsilon_{\mathrm{kill}},\\[4pt]
        0, & \zeta^{(\ell)}_{k} \le \epsilon_{\mathrm{kill}},
    \end{cases}
\end{equation}
with default threshold, $\epsilon_{\mathrm{kill}} = 0.5$.  
Units satisfying $\hat{\zeta}^{(\ell)}_{k}=0$ are removed entirely from the symbolic architecture, resulting in a compact and interpretable final representation.

Once the gates and masks are all sharpened, the resulting network contains exactly one primitive per edge with fixed affine parameters, one projection edge per unit, and only those units that survived pruning, yielding a sparse symbolic operator. This architecture reflects a discrete representation of the learned functional structure and can be used for symbolic analysis.

\subsection{Loss Function and Training Procedure}
\label{loss_sec}

Training minimizes a composite objective combining data fidelity or physics constraints with symbolic regularization. For supervised regression, the data loss is,
\begin{equation}
\mathcal{L}_{\mathrm{data}}
=\frac{1}{N_{\mathrm{tr}}}
\sum_{i=1}^{N_{\mathrm{tr}}}
\bigl(f_\theta(\boldsymbol{\xi}^{(i)}) - y^{(i)}\bigr)^2,
\end{equation}
for training pairs $\{(\boldsymbol{\xi}^{(i)},y^{(i)})\}_{i=1}^{N_{\mathrm{tr}}}$, where $\boldsymbol{\xi}^{(i)}$ denotes the $i$-th input vector, $y^{(i)}$ is its corresponding ground-truth value, and $N_{\mathrm{tr}}$ is the number of training samples. In PDE-constrained learning, the model $u_\theta$ approximates a solution of 
$\mathcal{N}[u]=0$ on a domain $\Omega$, with boundary condition 
$\mathcal{B}[u]=\mathcal{G}$ on $\partial\Omega$ and prescribed initial data.  
Let $\mathcal{S}_r$, $\mathcal{S}_b$, and $\mathcal{S}_0$ denote the sets of 
interior, boundary, and initial collocation points, respectively.  
The physics contributions are written as, 
\begin{equation}
\mathcal{L}_{\mathrm{PDE}}
=
\frac{1}{|\mathcal{S}_r|}
\sum_{\boldsymbol{\xi}\in\mathcal{S}_r}
\bigl|\mathcal{N}[u_\theta](\boldsymbol{\xi})\bigr|^2,
\qquad
\mathcal{L}_{\mathrm{BC}}
=
\frac{1}{|\mathcal{S}_b|}
\sum_{\boldsymbol{\xi}\in\mathcal{S}_b}
\bigl|\mathcal{B}[u_\theta](\boldsymbol{\xi})-\mathcal{G}(\boldsymbol{\xi})\bigr|^2,
\qquad
\mathcal{L}_{\mathrm{IC}}
=
\frac{1}{|\mathcal{S}_0|}
\sum_{\boldsymbol{\xi}\in\mathcal{S}_0}
\bigl|u_\theta(\boldsymbol{\xi},0)-u_0(\boldsymbol{\xi})\bigr|^2,
\end{equation}
and combined to form the complete physics loss as,
\begin{equation}
\mathcal{L}_{\mathrm{phys}}
=
\lambda_r\,\mathcal{L}_{\mathrm{PDE}}
+
\lambda_b\,\mathcal{L}_{\mathrm{BC}}
+
\lambda_0\,\mathcal{L}_{\mathrm{IC}},
\end{equation}
where $\lambda_r$, $\lambda_b$, and $\lambda_0$ are weighting coefficients.

Symbolic regularization on the other hand are used to act on the primitive gates $\boldsymbol{\alpha}_{ke}$. We apply the entropy and non-maximum-suppression (NMS) terms defined as, 
\begin{equation}
\mathcal{L}_{\mathrm{entropy}}
=\sum_{k=1}^{K_\ell}\sum_{e=1}^{E}\mathcal{H}(\boldsymbol{\alpha}_{ke}),\qquad
\mathcal{H}(\boldsymbol{\alpha}_{ke})=-\sum_{p=1}^{P}\alpha_{kep}\log\alpha_{kep},
\end{equation}
\begin{equation}
\mathcal{L}_{\mathrm{NMS}}
=\sum_{k=1}^{K_\ell}\sum_{1\le e_1<e_2\le E}\sum_{p=1}^{P}\alpha_{ke_1p}\alpha_{ke_2p},
\end{equation}
and combined as,
\begin{equation}
\mathcal{L}_{\mathrm{sel}}
=\lambda_{\mathrm{ent}}\mathcal{L}_{\mathrm{entropy}}
+\lambda_{\mathrm{nms}}\mathcal{L}_{\mathrm{NMS}},
\end{equation} 
 with $\lambda_{\mathrm{ent}}$ and $\lambda_{\mathrm{nms}}$ controlling sparsity and edge primitive diversity and  $\mathcal{H}(\boldsymbol{\alpha}_{ke})$ denoting the Shannon entropy \cite{shannon1948mathematical,faroughi2025neural} of the
primitive-selection distribution, penalizing diffuse mixtures and encouraging
sharp selections. 

Unit-level sparsity is also controlled through the continuous unit gates $\zeta^{(\ell)}_{k}$ for layer $\ell$ and unit $k$, producing,
\begin{equation}\label{unit_loss}
\mathcal{L}_{\mathrm{unit}}
=\sum_{\ell=1}^{L}\sum_{k=1}^{K_\ell}\zeta^{(\ell)}_{k}.
\end{equation}
It must be noted that the  additive penalty, Eq.~\eqref{unit_loss}, on unit gates may compete directly with the data or physics loss and can undesirably suppress a large number of units, especially when several units are jointly required to represent the underlying map. In such cases, it is preferable to replace the direct sum of gates with a soft budgeted sparsity that encourages each layer to maintain a target fraction of active units. Let $\rho\in(0,1)$ denote the desired proportion of active units in each layer. Defining the average gate value in layer $\ell$ as,
\begin{equation}
\bar{\zeta}^{(\ell)} = \frac{1}{K_\ell}\sum_{k=1}^{K_\ell} \zeta^{(\ell)}_{k},
\end{equation}
the budgeted unit sparsity penalty can be then written as,
\begin{equation}
\mathcal{L}_{\mathrm{unit}}
=
\sum_{\ell=1}^{L}
\bigl(\bar{\zeta}^{(\ell)} - \rho \bigr)^{2}.
\end{equation}
that prevents excessive deactivation while still allowing the model to learn a compact layerwise structure, with the target density $\rho$ controlling the desired sparsity level.  We finally apply a quadratic penalty on primitive output biases to prevent collapse to constant modes,
\begin{equation}
\mathcal{L}_{\mathrm{bias}}
=\sum_{k,e,p}\mathcal{B}_{kep}^{\,2}.
\end{equation}

The complete objective minimized during training thus is, 
\begin{equation}
\mathcal{L}(\theta)
=\lambda_{\mathrm{data}}\mathcal{L}_{\mathrm{data}}
+\mathcal{L}_{\mathrm{phys}}
+\lambda_{\mathrm{sel}}(t)\mathcal{L}_{\mathrm{sel}}
+\lambda_{\mathrm{unit}}\mathcal{L}_{\mathrm{unit}}
+\lambda_{\mathrm{bias}}\mathcal{L}_{\mathrm{bias}},
\end{equation}
where $\lambda_{\mathrm{data}}$, $\lambda_{\mathrm{bias}}$, and $\lambda_{\mathrm{unit}}$ are weighting coefficients and $\lambda_{\mathrm{sel}}(t)$ is the annealed (from zero) symbolic-regularization schedule. Training proceeds in two stages. Stage~I optimizes the relaxed primitive gates $\boldsymbol{\alpha}_{kep}$, the soft edge scores $\boldsymbol{S}_k$, and the annealed primitive temperature $\tau(t)$ that progressively sharpens the symbolic structure. When unit-level gates $\zeta^{(\ell)}_{k}$ are enabled, their continuous values are updated jointly to promote structural sparsity. Stage~II hardens all primitive gates and edge masks to their one-hot limits, optionally thresholds the unit gates to remove inactive units, and then refines the remaining continuous parameters with a second-order optimizer, such as L-BFGS \cite{faroughi2026KanRev,mostajeran2025scaled}.

For the dynamical system and physics-informed experiments, training stability becomes particularly important because the loss functions involve derivatives of the network outputs with respect to the inputs. These derivative-based objectives typically introduce additional stiffness into the optimization problem and make the training dynamics more sensitive to abrupt structural changes in the model. To mitigate this issue, the parameters controlling the symbolic gates, responsible for primitive selection and unit activation, are optimized more conservatively than the remaining network parameters. In practice, the gating variables are updated with a smaller effective learning rate and stronger regularization, allowing the continuous function parameters to adapt smoothly before discrete structural decisions are enforced. This separation of optimization time scales improves numerical stability and prevents premature gate saturation when learning dynamical laws or physics-constrained mappings. In addition, a gradually decaying learning-rate schedule is employed to further stabilize convergence as the symbolic structure sharpens during the later stages of training.

\section{Results and Discussion} \label{sec:results}

This section evaluates the performance of Symbolic--KAN across three learning paradigms: data-driven regression of multivariate functions, dynamical system identification, and physics-informed learning of partial differential equations. The first set of toy experiments examines whether the architecture can recover interpretable expressions directly from scattered data in spatiotemporal dimensions. The second set of experiments studies the capability of Symbolic--KAN to model nonlinear dynamical systems from data; in particular, we consider the Van der Pol equation and assess how accurately the learned model reproduces the underlying state dynamics. The third set of experiments investigates the ability of Symbolic--KAN to approximate solutions of partial differential equations (PDEs), testing both accuracy and symbolic interpretability. To quantify the predictive accuracy of the model, we use the relative error defined as,
\begin{equation}
    \mathcal{E}(F_{\boldsymbol{\theta}}) =
    \frac{\lVert F_{\boldsymbol{\theta}} - F \rVert}{\lVert F \rVert},
\end{equation}
where \(F_{\boldsymbol{\theta}}\) denotes the neural-network approximation of the true physical quantity \(F\). The notation \(\lVert \cdot \rVert\) refers to the $\mathcal{L}^2$ norm when \(F\) represents a function (e.g., a PDE solution field), and to the absolute value when \(F\) is a scalar parameter (e.g., a PDE coefficient in an inverse problem). 

\subsection{Data-driven Toy Experiments}\label{Sec:datadriven}
We first examine whether Symbolic--KAN can recover representations of
smooth nonlinear functions from scattered 
training samples. Each experiment uses a given number of training dataset
and a multilayer symbolic architecture with hardening applied at the end of phase I training.
Table~\ref{tab:function-regression} illustrates the performance of Symbolic--KAN on representative one-dimensional regression tasks. Note that in these toy examples, three edges are used per node. Table~\ref{tab:function-regression} reports the selected primitives for each node from the main library, along with the primary primitive selected after hardening. It also indicates whether each node is pruned. In the first experiment, where the target function is the simple polynomial \(F(x)=x^2\), the model successfully identifies the correct primitives \(x\)  and \(x^2\), in accordance with KART, within the symbolic structure. After the hardening stage, the resulting network retains the quadratic primitive while other candidates are pruned, leading to a highly accurate approximation with very small relative error. This behavior is also consistent with the performance characteristics of standard KAN architectures, which are known to efficiently represent low-degree polynomial relationships through their compositional structure. In the second experiment, as reported in Table~\ref{tab:function-regression}, the target function contains a richer nonlinear structure involving trigonometric components and a rational modulation term. Despite this increased complexity, the learned symbolic structure selects primitives such as \(\sin\), \(\cos\), and a Lorentz-type function (\(1/(1+x^2))\), which together form a meaningful approximation of the underlying expression. The selected primitives reflect key components of the target function, indicating that the proposed architecture is capable of capturing dominant functional patterns and constructing interpretable symbolic representations even when the true expression involves multiple interacting nonlinearities. This method could be used in future work as a pre-library selector for SINDy, addressing one of its key limitations~\cite{yonezawa2026sparse}.

\begin{table}[h!]
\centering
\caption{
Results of the data-driven regression tests for 1D functions (Section~\ref{Sec:datadriven}). For each target function, we report the network configuration $(L,K_{\ell},E)$, the number of training samples $N_{\mathrm{tr}}$, the relative prediction error $\mathcal{E}(F_{\boldsymbol{\theta}})$, and the discrete symbolic primitives selected by Symbolic--KAN before (reported in $|\cdot|$) and after the hardening stage, indicated by $\to$ symbol. Entries marked as \textit{alive} denote the unit retained in the final symbolic structure, whereas \textit{killed} indicates units pruned during training. 
}
\label{tab:function-regression}
\begin{tabularx}{\textwidth}{llll}
\toprule
\addlinespace
\textbf{Target Function} 
& $L,K_{\ell},E$ 
& $\mathcal{E}(F_{\boldsymbol{\theta}})$ 
& \textbf{Selected Primitives, Edge \& Unit} \\ \addlinespace 
\hline
\addlinespace
\begin{tabular}[c]{@{}l@{}}  
$F(x)=x^2$\\
$x\in[0,5]$ \\
$N_{tr}$ = 250 
\end{tabular}
& 2,2,3 
& $1.04\times10^{-5}$ 
& \begin{tabular}[c]{@{}l@{}}
$\ell=0,k=0$: |$x, x^2, x$| $\to$ [\boldsymbol{$x$}] (alive)\\
$\ell=0,k=1$: |$x^2$, cos, sin| $\to$ [sin] (killed)\\
$\ell=1,k=0$: |$x^2, x^2, x$| $\to$ [$x^2$] (killed)\\
$\ell=1,k=1$: |$x^2$, exp, cos| $\to$ [\boldsymbol{$x^2$}] (alive)
\end{tabular}
\\[8pt] 
\addlinespace
\hline
\addlinespace
\begin{tabular}[c]{@{}l@{}} 
$F(x)=\displaystyle{\frac{\sin(3x)}{1+x^2}+0.4\cos(5x)}$ \\
$x\in[0,5]$\\
$N_{tr}$ = 650 
\end{tabular}
& 2,3,3 
& $7.75\times10^{-3}$ 
& \begin{tabular}[c]{@{}l@{}}
$\ell=0,k=0$: |cos, $x$, $x^2$| $\to$ [\boldsymbol{$x^2$}] (alive)\\
$\ell=0,k=1$: |$x$, exp, sin| $\to$  [\textbf{sin}] (alive)\\
$\ell=1,k=0$: |lorentz, sin, $x$| $\to$ [\textbf{sin}] (alive)\\
$\ell=1,k=1$: |$x^2$, $x^2$, sin| $\to$ [\textbf{sin}] (alive) \\
$\ell=2,k=0$: |1, cos, exp| $\to$  [\textbf{cos}] (alive)\\
$\ell=2,k=1$: |sin, cos, lorentz| $\to$ [\textbf{lorentz}] (alive)

\end{tabular} \\[4pt]
\addlinespace
%

\bottomrule
\end{tabularx}
\end{table}

\begin{figure}[!h]
    \centering
    \includegraphics[width=1.0\linewidth]{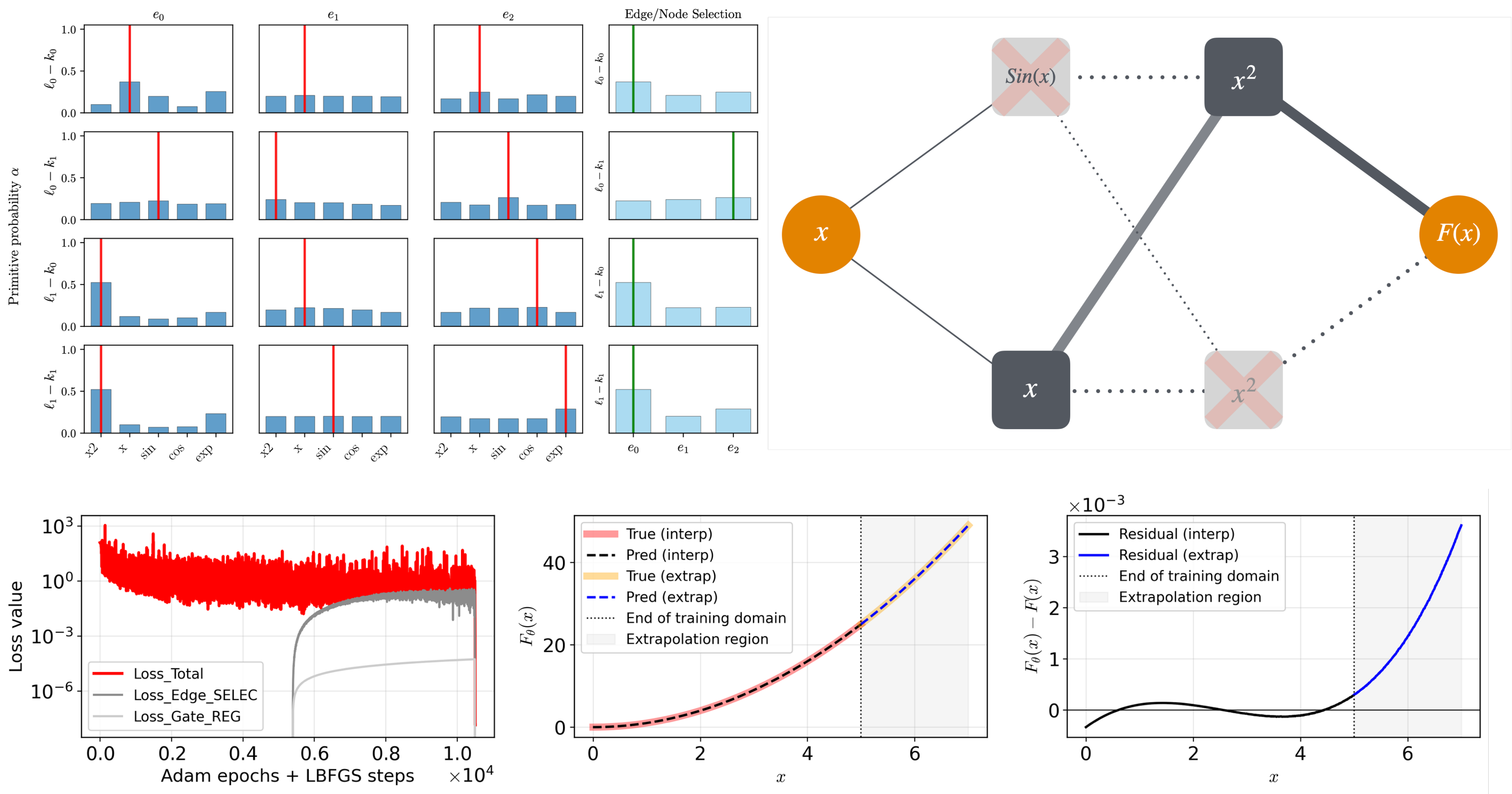}
    \caption{Symbolic--KAN reconstruction result for the target function \(F(x)=x^2\) in the data-driven regression experiment (Section~\ref{Sec:datadriven}). The upper-left panels illustrate how the model learns both the function's symbolic structure and its numerical behavior. Through gated optimization, only the relevant primitive corresponding to the linear and quadratic terms are retained, while redundant components are suppressed, yielding a compact interpretable form consistent with the ground truth. The final prediction closely matches the target function across the training region and extends smoothly beyond it, demonstrating accurate interpolation and, to some extent, extrapolation performance with consistently low residual error.}
    \label{fig:FirstExam0}
\end{figure}

Figure~\ref{fig:FirstExam0} illustrates the learning process and the resulting symbolic structure for the regression task with target function \(F(x)=x^2\). The network employs a two-layer (each with two units and three edges per units) Symbolic--KAN architecture in which each edges in a unit initially receives multiple candidate primitives per edge, while a gating mechanism selects the most relevant primitive during training. As shown in the upper-left panels, the gating variables progressively concentrate on a single dominant primitive for each active edge/unit, while competing candidates are suppressed and eventually pruned during the hardening stage. The resulting discrete structure is depicted on the top-right panel, where the network retains the linear \(x\) and quadratic primitive \(x^2\) for layer one and layer two, respectively, while removing unnecessary alternatives such as \(\sin(x)\) and \(x^2\) in each layer, yielding a compact symbolic representation consistent with the true target function and KART.
The lower panels provide further insight into the training dynamics and predictive behavior of the model. The loss curves demonstrate stable convergence of the optimization procedure as the gating and structural regularization terms guide the selection process. The prediction plots show that the learned model closely matches the ground-truth function within the training region, indicating accurate interpolation from the available samples. More importantly, the model also generalizes well beyond the observed domain: in the extrapolation region, where no training data are provided, the predicted curve continues to follow the correct quadratic trend. The residual plot confirms that the approximation error remains very small across both interpolation and extrapolation regimes.

\subsection{Dynamical System Experiments}
Understanding and identifying nonlinear dynamical systems from observational data is a fundamental problem in scientific computing and applied mathematics \cite{ADAMSINDy,howardsindy}. Many real-world physical, biological, and engineering processes are governed by nonlinear differential equations whose exact forms and parameters are often unknown \cite{Brunton2016}. In this context, dynamical system experiments provide a controlled yet challenging environment for evaluating the capability of data-driven frameworks to recover underlying governing laws. By focusing on systems with nonlinear interactions, state-dependent effects, and potentially non-integer power terms, we assess whether the proposed Symbolic--KAN method can simultaneously reconstruct the mathematical structure of the trajectories and accurately estimate unknown parameters. These experiments are designed not only to evaluate predictive accuracy but also to test structural identifiability, robustness to nonlinear complexity, and long-term stability in  trajectory reconstruction. We should note that in our experiments, Symbolic-KAN is approximating the solution trajectory and not the governing equation and the form of the governing equations are assumed to be known with unknown parameters.

\subsubsection{Van der Pol Equation}\label{Sec:vanderpol}

In this example, we evaluate the capability of Symbolic--KAN to discover governing equations from data by considering the Van der Pol oscillator, a classic model for nonlinear dynamical systems. Similar to~\cite{ADAMSINDy}, we modify the power to be a non-integer exponent. The oscillator is described by the following set of first-order differential equations,
\begin{equation}
    \begin{array}{ll}
\dot{x} = a\, y, \\[3mm]
\dot{y} = \mu (1 - x^{2.15})y - c\, x,
\end{array}
\end{equation}
where $a$, $\mu$, and $c$ are unknown parameters to be identified. The parameter $\mu > 0$ controls the strength of the nonlinear damping, which is responsible for the system's characteristic self-sustained oscillations. For small values of $\mu$, the system behaves like a simple harmonic oscillator. As $\mu$ increases, the nonlinearity intensifies, leading to the emergence of a stable limit cycle, where the amplitude and frequency of oscillations are inherently determined by the system's dynamics rather than initial conditions. This example presents a challenging test case for our proposed method due to the presence of a rational exponent ($x^{2.15}$) and the nonlinear damping term.
The presence of the rational exponent introduces a non-integer power nonlinearity, while the $\mu (1 - x^{2.15})y$ term captures state-dependent damping. These features create complex interactions that make it difficult to accurately extract the underlying structure from data.
The ground-truth system is simulated with the parameter values \(a = 1.0\), \(\mu = 0.01\), and \(c = 1.0\). To evaluate the proposed Symbolic--KAN framework, a high-accuracy reference solution is computed using an explicit Runge–Kutta method (RK45) with strict tolerance settings. Ground truth data are generated over the time interval \(t \in (0,T)\) with initial conditions \(x_0 = -2\) and \(y_0 = 0\), using a uniform time step of \(\Delta t = 0.01\).

\begin{table}[h!]
\centering
\caption{Results of the Van der Pol problem (Section~\ref{Sec:vanderpol}). All experiments use $|\mathcal{S}_r| = 10{,}000$ and $N_{\mathrm{tr}}=100$. The validation error $\mathcal{E}(u_{\boldsymbol{\theta}})$, representing the mean relative \(L^2\) error averaged over the state variables is reported along with the predicted parameters $a$, $\mu$, and $c$. In this experiment, we employ the function library $\{\sin, \cos, \exp, x, x^2\}$ and the network configuration $[L, K_{\ell}, E] = [4, 6, 3]$. The resulting symbolic primitives chosen after hardening are reported for each unit in each layer. Note that, in this example, the unit gates were disabled (i.e., no unit pruning). }
\label{tab:VdP-symbolic}
\begin{tabularx}{\textwidth}{XXXXXl}
\toprule
\textbf{Domain} 
& $\mathcal{E}(u_{\boldsymbol{\theta}})$ 
& \textbf{$a$} 
& \textbf{$\mu$} 
& \textbf{$c$} 
& \textbf{Selected Primitives} \\
\midrule
$t \in [0, 20]$ 
& $6.02\times10^{-4}$ 
& $1.0000$ 
& $0.0099$ 
& $0.9998$ 
&
\begin{tabular}[c]{@{}l@{}}
Layer 0:  $\to$ $[\sin, \sin, \sin, \sin, \sin, x]$ \\
Layer 1:  $\to$ $[x, \cos, x, \cos, \cos, \sin]$ \\
Layer 2:  $\to$ $[x, x, \sin, \sin, \cos, \cos]$ \\
Layer 3:  $\to$ $[\sin, \cos, \sin, \sin, x, \cos]$
\end{tabular} \\
\addlinespace
$t \in [0, 50]$ 
& $5.87\times10^{-3}$ 
& $0.9999$ 
& $0.0093$ 
& $0.9992$ 
&
\begin{tabular}[c]{@{}l@{}}
Layer 0:  $\to$ $[\cos, \cos, \cos, \sin, \cos, \sin]$ \\
Layer 1:  $\to$ $[\sin, \sin, \cos, x, \sin, \sin]$ \\
Layer 2:  $\to$ $[\sin, \sin, \cos, \sin, \cos, x]$ \\
Layer 3:  $\to$ $[\cos, \sin, x, \cos, \cos, \sin]$
\end{tabular} \\
\bottomrule
\end{tabularx}
\end{table}

Symbolic--KAN is trained on the generated time-series data to simultaneously reconstruct the governing structure and estimate the unknown coefficients in this inverse setting. To assess robustness with respect to the observation horizon and long-term nonlinear behavior, experiments are conducted for two different final times, \(T = 20\) and \(T = 50\). The corresponding quantitative results are presented in Table~\ref{tab:VdP-symbolic}. The reported error \( \mathcal{E}(u_{\boldsymbol{\theta}}) \) denotes the mean relative \(L^2\) error computed separately for each state variable (\(x_{\boldsymbol{\theta}}\) and \(y_{\boldsymbol{\theta}}\)) and then averaged, providing a scale-independent measure of trajectory accuracy.
As reported in Table~\ref{tab:VdP-symbolic}, Symbolic--KAN accurately recovers both the dynamical structure of the periodic solution trajectories and the unknown parameters. For \(T=20\), the identified coefficients are nearly exact: \(a\) and \(c\) match the true values up to four decimal digits, while the relative error in \(\mu\) is approximately \(1\%\). Even for the longer horizon \(T=50\), the parameter estimates remain highly accurate, with deviations below \(0.1\%\) for \(a\), about \(0.8\%\) for \(c\), and roughly \(7\%\) for \(\mu\), which is the most sensitive parameter governing nonlinear damping. Despite the increased temporal range and the presence of sustained oscillatory dynamics, the model maintains stable identification performance and low trajectory error.
These results demonstrate that Symbolic--KAN reliably extracts the governing equation parameters  directly from time-series data, achieving high predictive accuracy while preserving interpretability, even in the presence of nonlinear damping and non-integer power terms.

\begin{figure}[!h]
    \centering
    \includegraphics[width=1.0\linewidth]{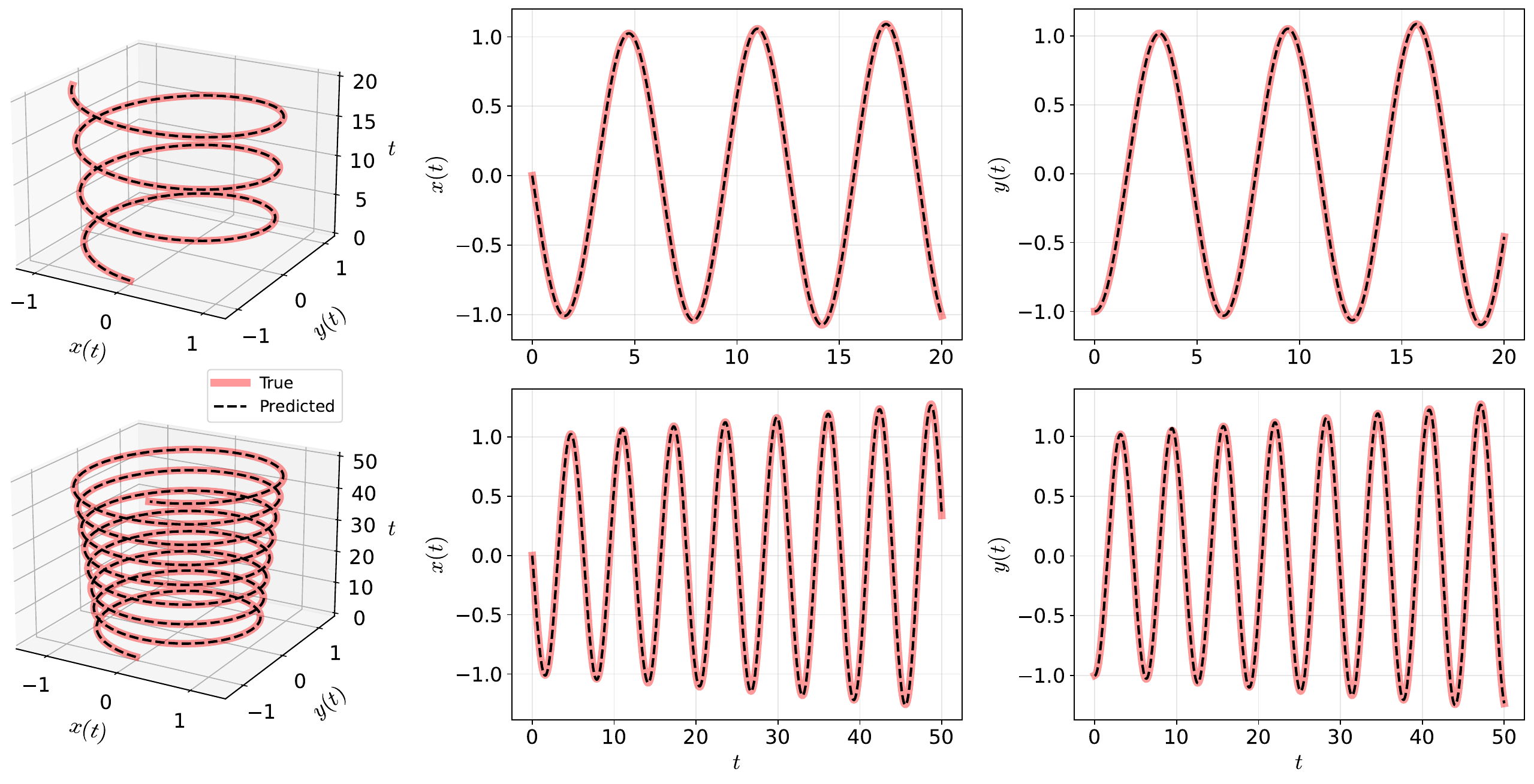}
    \caption{Comparison between the ground truth and Symbolic--KAN predicted solutions for the Van der Pol problem (Section~\ref{Sec:vanderpol}) over two observation horizons, $T=20$ (top row) and $T=50$ (bottom row). All experiments are conducted with $|\mathcal{S}_r| = 10{,}000$, $N_{\mathrm{tr}}=100$, and the network configuration $[L,K_{\ell},E]=[4,6,3]$.}
    \label{fig:VanDerPol_20and50}
\end{figure}

Figure~\ref{fig:VanDerPol_20and50} provides a comparison between the reference trajectories and the Symbolic--KAN predictions for both observation horizons. For \(T=20\), the predicted time evolutions of \(x(t)\) and \(y(t)\) are visually indistinguishable from the ground truth, with deviations remaining well below one tenth of a percent relative to the solution amplitude. The reconstructed phase trajectory in the three-dimensional \((x,y,t)\) space further confirms that the learned model accurately captures both the oscillatory behavior and the geometry of the limit cycle.
For the longer horizon \(T=50\), the agreement remains remarkably strong despite the extended nonlinear oscillations. The relative error stay at the level of roughly $< 1\%$, indicating that the identified model preserves both amplitude and phase over long time intervals. The overlap between true and predicted trajectories in phase space demonstrates that Symbolic--KAN not only fits the observed data but also reproduces the underlying dynamical structure over sustained evolution. The discrete primitives selected by Symbolic--KAN, as reported in Table~\ref{tab:VdP-symbolic}, are predominantly $\sin$, $\cos$, and $x$ across both time horizon, all of which can be directly linked to the intrinsic response of this particular dynamical system (i.e., the trigonometric primitives $\sin$ and $\cos$ naturally capture the oscillatory behavior of the system, while the linear term $x$ captures an approximately linear transient envelope over the observed time window, see Fig.~\ref{fig:VanDerPol_20and50}).

\begin{figure}[!h]
    \centering
    \includegraphics[width=0.950\linewidth]{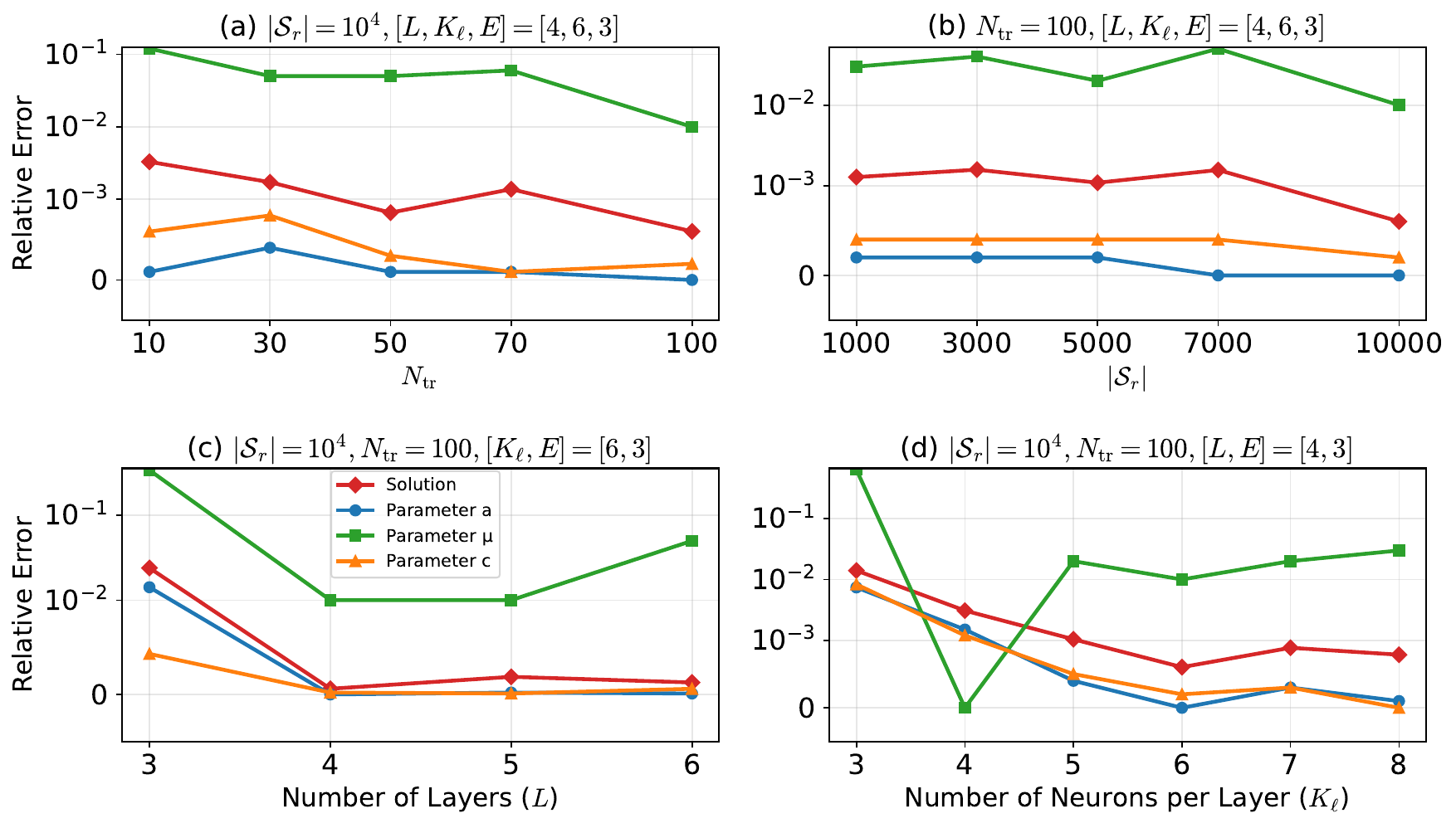}
    \caption{Performance assessment of the proposed Symbolic--KAN method for the Van der Pol problem (Section~\ref{Sec:vanderpol}). The relative errors of the reconstructed solution and the identified parameters \(a\), \(\mu\), and \(c\) are examined under systematic variations of the data and architectural hyperparameters. The figure illustrates the effect of changing the number of training samples \(N_{\mathrm{tr}}\) while keeping the number of interior collocation points and the network configuration fixed; varying the number of interior collocation points \(|\mathcal{S}_r|\) for a fixed training set and architecture; modifying the network depth \(L\) while maintaining the remaining hyperparameters fixed; and adjusting the number of neurons per layer, \(K_{\ell}\), with all other settings unchanged. The results highlight the robustness and stability of the proposed framework with respect to both data availability and network design choices.}
    \label{fig:VanDerPol_analysis}
\end{figure}

In addition, Fig.~\ref{fig:VanDerPol_analysis} examines the behavior of the proposed Symbolic--KAN method under different hyperparameter settings and data resolutions. 
With increasing numbers of training samples, the trajectory error and the parameter errors steadily decrease and reach the range of $10^{-3} \text{ and }10^{-4}$, demonstrating fast convergence as more observational data become available. A comparable behavior is observed when refining the interior collocation set: the errors remain small and stable, indicating that the method maintains accuracy without requiring excessive residual sampling. The absence of large oscillations or abrupt variations confirms the numerical robustness of the training procedure.
From the architectural perspective, even relatively compact networks provide accurate reconstructions. Although the nonlinear damping coefficient \(\mu\) is inherently more difficult to estimate due to its role in the state-dependent term, the method still recovers it with high precision, typically within a few percent and often substantially better. The remaining parameters \(a\) and \(c\) are identified with errors well below one percent across nearly all configurations.
Increasing the number of neurons per layer further improves stability and brings most errors below \(10^{-3}\). Therefore, Fig.~\ref{fig:VanDerPol_analysis} demonstrates that Symbolic--KAN delivers precise trajectory reconstruction and reliable parameter identification while remaining insensitive to reasonable variations in data density and network design.

\subsection{Physics-informed Experiments}

To assess the capability of Symbolic--KAN in physics-informed learning settings, we  consider resolving different partial differential equations for which the network is trained under the PDE residual, boundary conditions,  and initial conditions. 
After hardening, the symbolic structure is examined to determine whether the model  discovers meaningful discrete components in relation to  
the governing dynamics.

\subsubsection{Reaction--Diffusion Equation}\label{Exam.RD}

In this example, we assess the performance of the proposed Symbolic--KAN framework in an inverse problem setting by considering the following nonlinear reaction--diffusion equation \cite{mostajeran2025scaled},
\begin{equation}\label{Eq.RD}
\begin{array}{ll}
D\, u_{xx} + \kappa \tanh(u) = f, & x \in \Omega, \\[3mm]
u(x) = \sin^3(6 x), & x \in \partial \Omega,
\end{array}
\end{equation}
with \(\Omega = [-M, M]\). The parameter \(D\) denotes the diffusion coefficient, while \(\kappa\) controls the strength of the nonlinear reaction term.
The exact solution is prescribed as \(u(x) = \sin^3(6 x)\), and the corresponding source term \(f(x)\) is obtained analytically by substituting this expression into Eq.~\eqref{Eq.RD}. In the inverse setting, the goal is to reconstruct the underlying governing structure of the solution  and the corresponding coefficients in governing equations (e.g., in this case \(\kappa\)) from the produced data, thus establishing a rigorous benchmark for assessing both learning accuracy and stability. In this benchmark, the second-order diffusion operator, \(D\,u_{xx}\), introduces pronounced spatial variations, while the nonlinear reaction term, \(\kappa \tanh(u)\), yields strong state-dependent effects. The interaction between these two mechanisms produces a nontrivial solution structure, making reliable identification particularly demanding.
Through this example, we further examine the capability of Symbolic--KAN to handle nonlinear reaction--diffusion systems with nontrivial boundary conditions and to accurately reconstruct the underlying dynamics from data. The parameters of the problem are set to \(D = 10^{-2}\) and \(\kappa = 0.7\), and the spatial domain \(\Omega = [-M,M]\) is examined for \(M=2\) and \(M=4\). All models are trained with \(\lambda_{\text{PDE}} = 0.1\), a learning rate of \(5\times10^{-3}\), and 10,500 training epochs, using \(|\mathcal{S}_r|=5{,}000\) interior points and \(N_{\mathrm{tr}}=100\) measurements. 

Table~\ref{tab:RD-symbolic} presents a summary of the comparison results for this benchmark example.
Across both domains, the proposed Symbolic--KAN achieves high accuracy and consistently identifies parameters correctly. On the smaller interval \([-2,2]\), the validation error remains below \(0.1\%\), and the recovered reaction coefficient differs from its true value by less than \(0.1\%\). When the domain is extended to \([-4,4]\), the problem becomes more demanding due to increased spatial variation; nevertheless, the method maintains a validation error at the level of roughly \(1\%\), while estimating \(\kappa\) with a relative error of about \(0.2\%\). These findings demonstrate that the method remains stable when the domain is enlarged, without requiring any extra scaling or stabilization techniques, as discussed in depth in \cite{mostajeran2025scaled}. An important outcome concerns symbolic discovery. Using an enriched library, the hardened architecture consistently selects trigonometric primitives as dominant components across layers, with their prevalence increasing as the spatial domain is enlarged. This trend is physically meaningful: as the domain expands, the underlying functional structure becomes more fully expressed, and the model correspondingly favors the appropriate basis functions. This behavior aligns with the analytical form of the ground-truth solution $u(x)=\sin^3(6x)$, indicating that Symbolic--KAN progressively uncovers the correct functional representation rather than merely fitting data numerically, which is a key improvement toward the elusive goal of mechanistic learning in deep neural networks. The resulting symbolic structure further supports the claim that the learned representation generalizes across spatial scales.

\begin{table}[t]
\centering
\caption{Results for the 1D reaction--diffusion problem (Section~\ref{Exam.RD}). All experiments are conducted with \( |\mathcal{S}_r| = 5{,}000 \) interior collocation points and \( N_{\mathrm{tr}} = 100 \) training samples. The table reports, for each method, the network configuration \((L, K_{\ell}, E)\), the validation error \( \mathcal{E}(u_{\boldsymbol{\theta}})\), and the identified reaction coefficient \( \kappa \). For the Symbolic--KAN model, the selected symbolic primitives after the hardening stage are also listed. The candidate library used for symbolic discovery is \( \{\sin, \cos, \exp, x, x^2\}\).
For baseline models, \( k_{\mathrm{Cheb}} \) denotes the degree of the Chebyshev polynomial expansion used in cPIKAN. Note that, in this example, the unit gates were disabled (i.e., no unit pruning).}  %
\label{tab:RD-symbolic}
\begin{tabularx}{\textwidth}{XXXXXl}
\toprule
\textbf{Domain} & \textbf{Method} & $L,K_{\ell},E$ & $\mathcal{E}(u_{\boldsymbol{\theta}})$ & $\kappa$ & \textbf{Selected Primitives} \\
\midrule
$x \in [-2,2]$ & Symbolic--KAN & $4,6,3$ & $5.93\times10^{-4}$ & $0.6994$ &
\begin{tabular}[c]{@{}l@{}}
Layer 0: $\to$ $[\cos, \exp, \exp, x^2, x^2, x^2]$ \\
Layer 1: $\to$ $[\cos, x, \sin, \sin, \sin, x^2]$ \\
Layer 2: $\to$ $[\sin, x, \sin, \sin, x, \cos]$ \\
Layer 3: $\to$ $[\cos, \cos, x^2, \cos, x, \cos]$
\end{tabular} \\
\addlinespace
$x \in [-4,4]$ & Symbolic--KAN & $4,6,3$ & $9.37\times10^{-3}$ & $0.6985$ &
\begin{tabular}[c]{@{}l@{}}
Layer 0:  $\to$ $[\sin, x, \cos, \sin, \sin, x]$ \\
Layer 1: $\to$ $[\sin, \sin, \sin, x^2, \sin, \sin]$ \\
Layer 2: $\to$ $[\cos, \sin, \cos, \cos, \sin, \sin]$ \\
Layer 3: $\to$  $[\sin, \cos, \sin, \cos, \sin, \sin]$
\end{tabular} \\
\midrule
\midrule
\multicolumn{6}{l}{Comparison with baseline models (no symbolic discovery)} \\
\midrule
\textbf{Domain} & \textbf{Method} & $L,K_{\ell}, k_{\mathrm{Cheb}}$ & $\mathcal{E}(u_{\boldsymbol{\theta}})$ & $\kappa$ & \\
\midrule
$x \in [-2,2]$ & cPIKAN & $4,6,3$ & $8.25\times10^{-4}$ & $0.6998$ & -- \\
\addlinespace
 & PINN & $4,12,-$ & $1.50\times10^{-1}$ & $0.6899$ & -- \\
\addlinespace
$x \in [-4,4]$ & cPIKAN & $4,6,3$ & $2.07\times10^{-1}$ & $0.6611$ & -- \\
\addlinespace
 & PINN & $4,12,-$ & $2.15\times10^{-1}$ & $0.6809$ & -- \\
\bottomrule
\end{tabularx}
\end{table}

\begin{figure}[!h]
    \centering
    \includegraphics[width=1.0\linewidth]{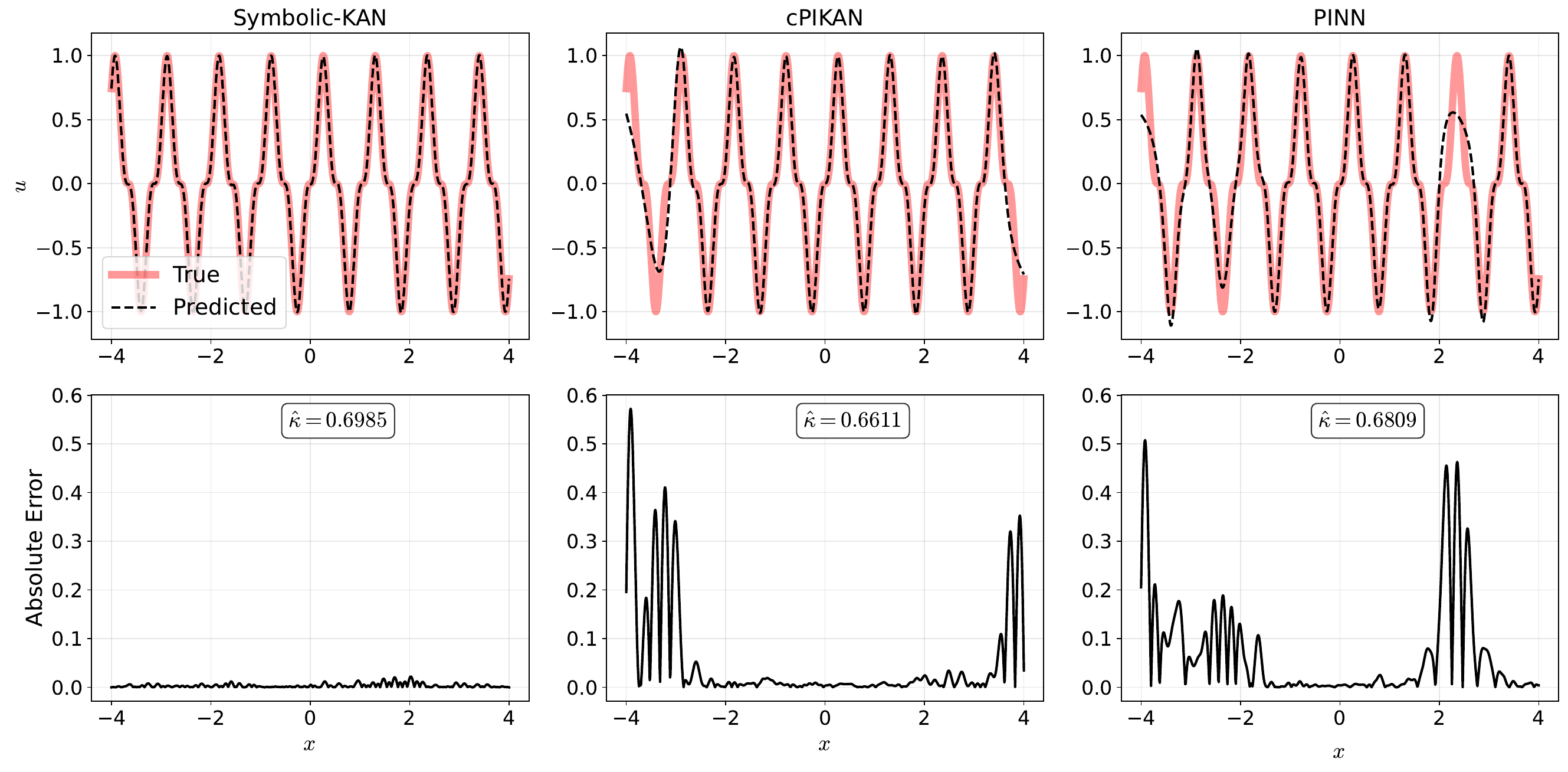}
    \caption{Comparative evaluation of Symbolic--KAN, cPIKAN, and PINN for the 1D reaction--diffusion problem (Section~\ref{Exam.RD}) on the extended domain \(x \in [-4,4]\). The predicted reaction coefficient \(\hat{\kappa}\) is also reported in each case. All network configurations and training settings are consistent with those listed in Table~\ref{tab:RD-symbolic}. The comparison highlights the accuracy of the proposed Symbolic--KAN framework in both solution reconstruction and parameter identification relative to vanilla cPIKAN and PINN.
}
    \label{fig:rd_4}
\end{figure}

The comparison with baseline models further clarifies the benefit of embedded symbolic discovery. On a domain with \([-2,2]\), Symbolic--KAN and cPIKAN (Chebyshev Physics-Informed KAN \cite{mostajeran2025scaled}) achieve comparable accuracy, with validation errors differing by roughly \(28\%\) in relative terms, while both estimate \(\kappa\) with errors below \(0.05\%\). However, when the domain expands to \([-4,4]\), their behavior diverges significantly. Symbolic--KAN reduces the validation error by approximately \(95\%\) compared to cPIKAN and estimates \(\kappa\) more than five times more accurately. The contrast with a standard PINN (Physics-Informed Neural Network using \(\tanh\) activation function \cite{mostajeran2025scaled}) is even more pronounced: for both domains, the validation error of Symbolic--KAN is over \(99\%\) lower, and the maximum relative error in \(\kappa\) remains below \(0.3\%\), whereas the PINN exhibits parameter errors on the order of several percent. Therefore, the reaction--diffusion experiment confirms that Symbolic--KAN effectively integrates high numerical accuracy with interpretable symbolic structure. It preserves stability under domain extension, maintains precise parameter recovery, and provides clear advantages over purely numerical counterparts, establishing it as a robust and interpretable framework for inverse problems in scientific machine learning.

Figure \ref{fig:rd_4} provides a detailed visual comparison on the extended domain \([-4,4]\), further emphasizing the robustness of the proposed approach. In terms of solution accuracy, Symbolic--KAN achieves a large reduction in maximum absolute error, decreasing the error by approximately 96.7\% relative to cPIKAN and 96.0\% relative to PINN. This substantial improvement highlights its superior capability in capturing the spatial structure of the solution over larger domains.
A similar pattern is observed in parameter identification. The proposed framework reduces the relative error in estimating the reaction coefficient by about 96.2\% compared to cPIKAN and 92.3\% compared to PINN. These results clearly demonstrate that, under domain enlargement, Symbolic--KAN consistently delivers  higher reconstruction accuracy and significantly more precise parameter recovery than the competing approaches, confirming its stability and robustness in more demanding settings.

\subsubsection{Laplace Equation}\label{sec:lapeq}
As the second PDE benchmark, we consider the two-dimensional Laplace equation on a 
rectangular domain,
\begin{equation}
\label{eq:laplace}
\nabla^{2}u(\boldsymbol{\xi})
=
u_{xx}(\boldsymbol{\xi}) + u_{yy}(\boldsymbol{\xi})
= 0,
\qquad 
\boldsymbol{\xi}=(x,y)\in\Omega=[0,1]\times[0,1].
\end{equation}
that admits a wide class of analytical solutions determined entirely 
by boundary data. For this study, we prescribe Dirichlet conditions on all 
boundaries such that the analytical solution is determined in closed form,
\begin{equation}
u(x,y)
=
\sin(\pi x)\,\sinh(\pi y),
\end{equation}
for which
\(
\nabla^{2}u = 0
\)
is satisfied identically. 
The boundary conditions are then taken directly from the exact solution.
For this benchmark example, the Symbolic--KAN model is trained using the physics-informed loss 
$\mathcal{L}_{\mathrm{phys}}$ defined in Section~\ref{loss_sec}
with residual points enforcing
\(
\nabla^{2}u_\theta(\boldsymbol{\xi})=0
\)
and other regularization terms. All experiments are conducted with \( |\mathcal{S}_r| = 10{,}000 \) interior collocation points and \( |\mathcal{S}_b| = 400 \) boundary collocation points  sampled uniformly inside the domain and on its boundaries. Moreover, $N_{\mathrm{tr}} = 0$ because this experiment does not perform any parameter estimation in inverse mode.

\begin{table}[h!]
\centering
\caption{Results for the Laplace problem (Section~\ref{sec:lapeq}). All experiments are conducted with \( |\mathcal{S}_r| = 10{,}000 \) interior collocation points and \( |\mathcal{S}_b| = 400 \) boundary collocation points. For each method, the network configuration \((L, K_{\ell}, E)\), and the validation error \( \mathcal{E}(u_{\boldsymbol{\theta}})\) are reported. For the Symbolic--KAN model, the selected symbolic primitives after the hardening stage are also listed. The candidate library used for symbolic discovery is \( \{1, x, x^2, \sin, \cos, \sinh, \cosh, \exp \} \).
For baseline models, $k_{\mathrm{Cheb}}$ denotes the degree of the Chebyshev polynomial expansion used in cPIKAN. Note that, in this example, the unit gates were disabled (i.e., no unit pruning).}  
\label{tab:lap-symbolic}
\begin{tabularx}{\textwidth}{XXXc}
\toprule
 \textbf{Method} & $L,K_{\ell},E$ & $\mathcal{E}(u_{\boldsymbol{\theta}})$ &  \textbf{Selected Primitives} \\
\midrule
\addlinespace
  Symbolic--KAN & $4,6,3$ & $1.11\times10^{-3}$ &  
\begin{tabular}[c]{@{}l@{}}
Layer 0: $\to$ $[\cos, \sin, \cosh, \sinh, \cosh, x^2]$ \\
Layer 1: $\to$  $[\sin, x, x,  \sin, x, \sin]$ \\
Layer 2: $\to$  $[\text{const}, \sin, \sin, \sinh, \cos, \cos]$ \\
Layer 3: $\to$ $[\sinh, \cosh, \sinh, \sin, \cosh, \cosh]$
\end{tabular} \\
\midrule
\midrule
\multicolumn{4}{l}{Comparison with baseline models (no symbolic discovery)} \\
\midrule
\textbf{Method} & $L,K_{\ell},k_{\mathrm{Cheb}}$ & $\mathcal{E}(u_{\boldsymbol{\theta}})$ &  \\
\midrule
  cPIKAN & $4,6,3$ & $8.76\times10^{-3}$   & -- \\
\addlinespace
  PINN & $4,18, -$ & $2.71\times10^{-3}$ &   -- \\
\addlinespace
\bottomrule
\end{tabularx}
\end{table}

The results in Table~\ref{tab:lap-symbolic} show that the Symbolic--KAN model selects primitives that are consistent with the functional structure of the exact solution. The analytical solution of the problem is \(u(x,y)=\sin(\pi x)\sinh(\pi y)\), which is composed of trigonometric and hyperbolic functions. Interestingly, after the hardening stage the discovered primitives mainly include \(\sin\), \(\cos\), \(\sinh\), and \(\cosh\), together with simple identity mappings. This indicates that the symbolic discovery mechanism successfully identifies basis functions that belong to the same functional family as the true solution, even though the exact expression itself is not explicitly provided to the model. Such behavior suggests that the method is capable of uncovering meaningful structural components of the underlying solution. Table~\ref{tab:lap-symbolic} also highlights  the quantitative advantage of the proposed approach when compared with the two baseline methods. In terms of validation error, Symbolic--KAN achieves a substantial improvement of approximately 59\% over the standard PINN (using \(\tanh\) activation function) and about 87\% over cPIKAN. This consistent reduction in error indicates that, beyond interpretability, incorporating symbolic discovery leads to a more efficient representation of the solution space and improved generalization performance for this elliptic problem.

\begin{figure}[!h]
    \centering
    \includegraphics[width=1.0\linewidth]{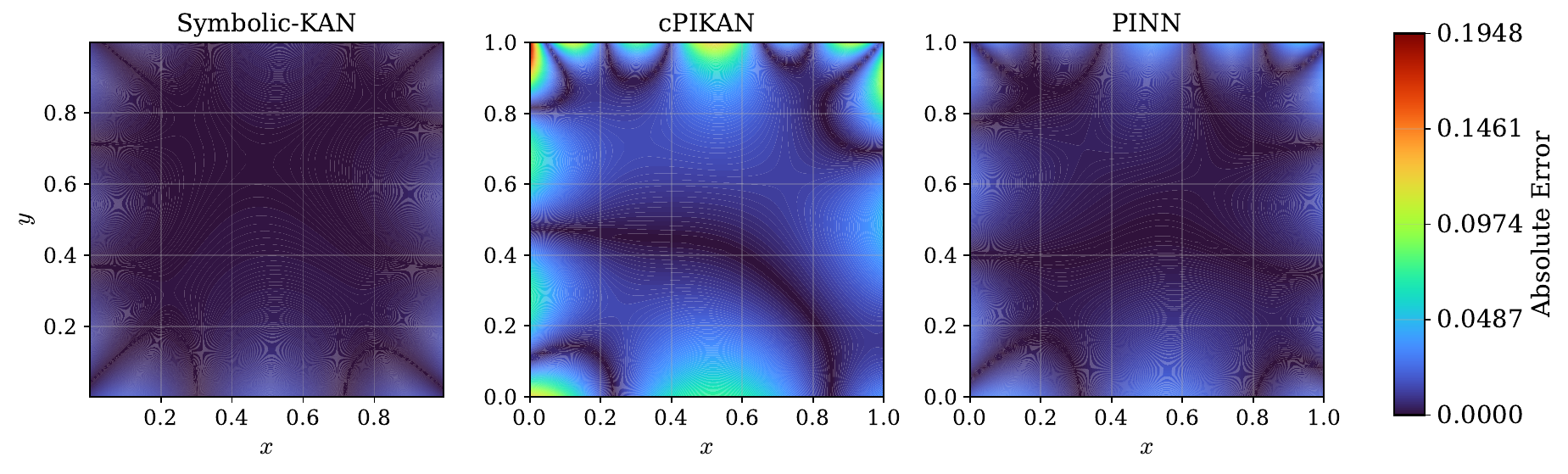}
    \caption{Comparative evaluation of Symbolic--KAN, cPIKAN, and PINN for Laplace equation (Section~\ref{sec:lapeq}).  All network configurations and training settings are  listed in Table~\ref{tab:lap-symbolic}. The comparison highlights the accuracy of the proposed Symbolic--KAN framework relative to vanilla cPIKAN and PINN.
}
    \label{fig:lap}
\end{figure}

Figure~\ref{fig:lap} provides a more detailed demonstration of the accuracy of the proposed Symbolic--KAN model by comparing the absolute errors of all three models listed in Table~\ref{tab:lap-symbolic}. Among them, the cPIKAN model shows the largest deviation from the reference solution, this error is substantially reduced by the PINN approach, and then further diminished by Symbolic--KAN, clearly indicating an improvement in solution quality achieved with the proposed Symbolic--KAN framework. In particular, the maximum absolute error obtained by Symbolic--KAN is substantially smaller, corresponding to roughly a 70\% reduction compared with PINN and about a 92\% reduction compared with cPIKAN. This comparison confirms that Symbolic--KAN not only captures the correct functional structure but also provides a noticeably more accurate approximation of the solution across the domain.



\section{Conclusion}\label{sec:conclusion}
In this work, we presented Symbolic Kolmogorov--Arnold Networks (Symbolic--KANs) as a unified framework for scalable learning and interpretable symbolic discovery in scientific machine learning. By embedding discrete symbolic structure within a trainable architecture, the proposed approach connects the expressive power of neural networks with the transparency of symbolic models, enabling the recovery of compact analytic representations directly from data and governing constraints. Across a range of settings, including data-driven regression, inverse dynamical systems such as the Van der Pol oscillator, and physics-informed learning of equations such as the reaction--diffusion systems and the Laplace problem, Symbolic--KAN consistently identifies meaningful primitive structures while maintaining strong predictive accuracy compared to baseline methods. In these cases, the learned representations not only match the target solutions but also recover the underlying analytical structure, highlighting the ability of the framework to extract mechanistic insight rather than purely predictive mappings.

More broadly, Symbolic--KAN provides a practical pathway toward mechanistic machine learning, where models are both trainable at scale and amenable to symbolic inspection. By enabling the discovery and refinement of interpretable functional primitives, the framework supports downstream tasks such as equation discovery, model reduction, and scientific interpretation. These results point toward a new class of learning frameworks that unify expressivity, sparsity, and interpretability, advancing the development of robust and interpretable neural solvers in settings where the governing structure must be inferred from data.

\section{Data and Code Availability}
All code and data will be released upon publication.


\section{Conflict of Interest}
The authors declare no conflict of interests.

\def\mybibdoicolor{\color{black}}
\newcommand*{\doi}[1]{\href{\detokenize{#1}} {\raggedright\mybibdoicolor{DOI: \detokenize{#1}}}}

\bibliographystyle{elsarticle-num}
\bibliography{references.bib}

@inproceedings{kolmogorov1957representations,
  title={On the representations of continuous functions of many variables by superposition of continuous functions of one variable and addition},
  author={Kolmogorov, Andrei Nikolaevich},
  booktitle={Dokl. Akad. Nauk USSR},
  volume={114},
  pages={953--956},
  year={1957},
  doi = {},
  url = {}
}

@article{howardsindy,
      title={SINDy-KANs: Sparse identification of non-linear dynamics through Kolmogorov-Arnold networks}, 
      author={Amanda A. Howard and Nicholas Zolman and Bruno Jacob and Steven L. Brunton and Panos Stinis},
      year={2026},
      eprint={2603.18548},
      archivePrefix={arXiv},
      primaryClass={cs.LG},
      url={https://arxiv.org/abs/2603.18548}, 
}

@article{Arnold1958,
  title={On the representation of functions of several variables by superpositions of functions of
fewer variables},
  author={Arnold, V.},
  journal={Mat. Prosvesh.},
  volume={3},
  number={},
  pages={41–61},
  year={1958},
  publisher={},
  doi = {},
  url = {}
}

@book{kolmogorov1961representation,
  title={On the representation of continuous functions of several variables by superpositions of continuous functions of a smaller number of variables},
  author={Kolmogorov, Andre{\u\i} Nikolaevich},
  year={1961},
  publisher={American Mathematical Society},
doi = {},
url = {}
}

@article{strypsteen2021end,
  title={End-to-end learnable EEG channel selection for deep neural networks with Gumbel-softmax},
  author={Strypsteen, Thomas and Bertrand, Alexander},
  journal={Journal of Neural Engineering},
  volume={18},
  number={4},
  pages={0460a9},
  year={2021},
doi = {https://doi.org/10.1088/1741-2552/ac115d},
  publisher={IOP Publishing}
}

@article{Wang2020NTKPinns,
  title={When and why PINNs fail to train: A neural tangent kernel perspective},
  author={Wang, Sifan and Yu, Xinling and Perdikaris, Paris},
  journal={Journal of Computational Physics},
  volume={449},
  pages={110768},
  year={2022},
  publisher={Elsevier},
  doi = {https://doi.org/10.1016/j.jcp.2021.110768},
  url = {https://www.sciencedirect.com/science/article/abs/pii/S002199912100663X}
}

@article{Li2020FNO,
  title={Fourier neural operator for parametric partial differential equations},
  author={Li, Zongyi and Kovachki, Nikola and Azizzadenesheli, Kamyar and Liu, Burigede and Bhattacharya, Kaushik and Stuart, Andrew and Anandkumar, Anima},
  journal={arXiv preprint arXiv:2010.08895},
  year={2020},
  doi = {https://doi.org/10.48550/arXiv.2010.08895},
  url = {https://arxiv.org/abs/2010.08895}
}

@article{abueidda2025deepokan,
  title={Deep{OKAN}: Deep operator network based on {K}olmogorov {A}rnold networks for mechanics problems},
  author={Abueidda, Diab W and Pantidis, Panos and Mobasher, Mostafa E},
  journal={Computer Methods in Applied Mechanics and Engineering},
  volume={436},
  pages={117699},
  year={2025},
  publisher={Elsevier},
  doi = {https://doi.org/10.1016/j.cma.2024.117699},
  url = {https://www.sciencedirect.com/science/article/pii/S0045782524009538}
}

@article{mostajeran2025scaled,
title = {Scaled-c{PIKAN}s: Spatial variable and residual scaling in {C}hebyshev-based physics-informed {K}olmogorov-{A}rnold networks},
journal = {Journal of Computational Physics},
volume = {537},
pages = {114116},
year = {2025},
issn = {0021-9991},
doi = {https://doi.org/10.1016/j.jcp.2025.114116},
url = {https://www.sciencedirect.com/science/article/pii/S0021999125003997},
author = {Farinaz Mostajeran and Salah A. Faroughi}
}

@article{schmidt2021kolmogorov,
  title={The {K}olmogorov--{A}rnold representation theorem revisited},
  author={Schmidt-Hieber, Johannes},
  journal={Neural networks},
  volume={137},
  pages={119--126},
  year={2021},
  publisher={Elsevier},
doi = {https://doi.org/10.1016/j.neunet.2021.01.020},
url = {https://www.sciencedirect.com/science/article/pii/S0893608021000289}
}

@article{lai2021kolmogorov,
  title={The kolmogorov superposition theorem can break the curse of dimensionality when approximating high dimensional functions},
  author={Lai, Ming-Jun and Shen, Zhaiming},
  journal={arXiv preprint arXiv:2112.09963},
  year={2021},
doi = {https://doi.org/10.48550/arXiv.2112.09963},
url = {https://arxiv.org/abs/2112.09963}
}

@article{fakhoury2022exsplinet,
  title={Ex{S}pli{N}et: An interpretable and expressive spline-based neural network},
  author={Fakhoury, Daniele and Fakhoury, Emanuele and Speleers, Hendrik},
  journal={Neural Networks},
  volume={152},
  pages={332--346},
  year={2022},
  publisher={Elsevier},
doi = {https://doi.org/10.1016/j.neunet.2022.04.029},
url = {https://www.sciencedirect.com/science/article/abs/pii/S0893608022001617}
}

@article{he2023optimal,
  title={On the optimal expressive power of relu dnns and its application in approximation with kolmogorov superposition theorem},
  author={He, Juncai},
  journal={arXiv preprint arXiv:2308.05509},
  year={2023},
doi = {https://doi.org/10.1109/TNNLS.2024.3514126},
url = {https://ieeexplore.ieee.org/abstract/document/10804852}
}

@article{li2024hkan,
  title={{HKAN}: A Hybrid {K}olmogorov--{A}rnold Network for Robust Fabric Defect Segmentation},
  author={Li, Min and Ye, Pei and Cui, Shuqin and Zhu, Ping and Liu, Junping},
  journal={Sensors},
  volume={24},
  number={24},
  pages={8181},
  year={2024},
  publisher={MDPI},
doi = {https://doi.org/10.3390/s24248181},
url = {https://www.mdpi.com/1424-8220/24/24/8181}
}

@article{liu2024kan,
  title={Kan: Kolmogorov-arnold networks},
  author={Liu, Ziming and Wang, Yixuan and Vaidya, Sachin and Ruehle, Fabian and Halverson, James and Solja{\v{c}}i{\'c}, Marin and Hou, Thomas Y and Tegmark, Max},
  journal={arXiv preprint arXiv:2404.19756},
  year={2024},
doi = {https://doi.org/10.48550/arXiv.2404.19756},
url = {https://arxiv.org/abs/2404.19756}
}

@article{li2024kolmogorov,
  title={Kolmogorov-arnold networks are radial basis function networks},
  author={Li, Ziyao},
  journal={arXiv preprint arXiv:2405.06721},
  year={2024},
doi = {https://doi.org/10.48550/arXiv.2405.06721},
url = {https://arxiv.org/abs/2405.06721}
}

@article{ta2024bsrbf,
  title={{BSRBF}-{KAN}: {A} combination of {B}-splines and {R}adial {B}asis {F}unctions in {K}olmogorov-{A}rnold {N}etworks},
  author={Ta, Hoang-Thang},
  journal={arXiv preprint arXiv:2406.11173},
  year={2024},
doi = {https://doi.org/10.1007/978-981-96-4288-5_1},
url = {https://link.springer.com/chapter/10.1007/978-981-96-4288-5_1}
}

@article{liu2024kan2,
  title={{KAN} 2.0: Kolmogorov-{A}rnold networks meet science},
  author={Liu, Ziming and Ma, Pingchuan and Wang, Yixuan and Matusik, Wojciech and Tegmark, Max},
  journal={arXiv preprint arXiv:2408.10205},
  year={2024},
doi = {https://doi.org/10.48550/arXiv.2408.10205},
url = {https://arxiv.org/abs/2408.10205}
}

@article{mostajeran2026epi,
  title={Elastoplasticity informed kolmogorov--arnold networks using chebyshev polynomials},
  author={Mostajeran, Farinaz and Faroughi, Salah A},
  journal={International Journal for Numerical and Analytical Methods in Geomechanics},
volume={},
  number={},
  pages={},
  year={2026},
  publisher={Wiley},
doi = {https://doi.org/10.1002/nag.70283},
url = {}
}

@article{ta2025prkan,
  title={{PRKAN}: Parameter-{R}educed {K}olmogorov-{A}rnold {N}etworks},
  author={Ta, Hoang-Thang and Thai, Duy-Quy and Tran, Anh and Sidorov, Grigori and Gelbukh, Alexander},
  journal={arXiv preprint arXiv:2501.07032},
  year={2025},
doi = {https://doi.org/10.48550/arXiv.2501.07032},
url = {https://arxiv.org/abs/2501.07032}
}

@article{koenig2024kan,
  title={{KAN}-{ODE}s: Kolmogorov--{A}rnold network ordinary differential equations for learning dynamical systems and hidden physics},
  author={Koenig, Benjamin C and Kim, Suyong and Deng, Sili},
  journal={Computer Methods in Applied Mechanics and Engineering},
  volume={432},
  pages={117397},
  year={2024},
  publisher={Elsevier},
doi = {https://doi.org/10.1016/j.cma.2024.117397},
url = {https://www.sciencedirect.com/science/article/abs/pii/S0045782524006522}
}

@article{schmidt2020nonparametric,
author = {Johannes Schmidt-Hieber},
title = {{Nonparametric regression using deep neural networks with {R}e{LU} activation function}},
volume = {48},
journal = {The Annals of Statistics},
number = {4},
publisher = {Institute of Mathematical Statistics},
pages = {1875 -- 1897},
year = {2020},
doi = {10.1214/19-AOS1875},
URL = {https://doi.org/10.1214/19-AOS1875}
}

@article{wang2025efkan,
  title={{EFKAN}: A {KAN}-Integrated Neural Operator For Efficient Magnetotelluric Forward Modeling},
  author={Wang, Feng and Qiu, Hong and Huang, Yingying and Gu, Xiaozhe and Wang, Renfang and Yang, Bo},
  journal={arXiv preprint arXiv:2502.02195},
  year={2025},
doi = {https://doi.org/10.48550/arXiv.2502.02195},
url = {https://arxiv.org/abs/2502.02195}
}

@article{liang2026wkpnet,
  title={WKPNet: A novel wavelet-KAN-POLA network for medical image segmentation},
  author={Liang, Pengchen and Zeng, Quanhong and Liang, Bocheng and Huang, Haishan and Zhang, Yudong and Pu, Bin and Chen, Jianguo},
  journal={Biomedical Signal Processing and Control},
  volume={113},
doi = {https://doi.org/10.1016/j.bspc.2025.108988},
  pages={108988},
  year={2026},
  publisher={Elsevier}
}

@article{jacob2024spikans,
  title={{SPIKAN}s: Separable physics-informed kolmogorov-arnold networks},
  author={Jacob, Bruno and Howard, Amanda and Stinis, Panos},
  journal={Machine Learning: Science and Technology},
  year={2024},
doi = {https://doi.org/10.1088/2632-2153/ae05af},
url = {https://iopscience.iop.org/article/10.1088/2632-2153/ae05af}
}

@article{toscano2025kkans,
  title={{KKAN}s: K\.{u}rkov\'{a}-{K}olmogorov-{A}rnold networks and their learning dynamics},
  author={Toscano, Juan Diego and Wang, Li-Lian and Karniadakis, George Em},
  journal={Neural Networks},
  pages={107831},
  year={2025},
  publisher={Elsevier},
doi = {https://doi.org/10.1016/j.neunet.2025.107831},
url = {https://www.sciencedirect.com/science/article/pii/S0893608025007117}
}

@article{daryakenari2025representation,
  title={Representation Meets Optimization: Training PINNs and PIKANs for Gray-Box Discovery in Systems Pharmacology},
  author={Daryakenari, Nazanin Ahmadi and Shukla, Khemraj and Karniadakis, George Em},
  journal={arXiv preprint arXiv:2504.07379},
  year={2025},
doi = {https://doi.org/10.48550/arXiv.2504.07379},
url = {https://arxiv.org/abs/2504.07379}
}

@article{chen2024sckansformer,
  title={{SCK}ansformer: Fine-grained classification of bone marrow cells via kansformer backbone and hierarchical attention mechanisms},
  author={Chen, Yifei and Zhu, Zhu and Zhu, Shenghao and Qiu, Linwei and Zou, Binfeng and Jia, Fan and Zhu, Yunpeng and Zhang, Chenyan and Fang, Zhaojie and Qin, Feiwei and others},
  journal={IEEE Journal of Biomedical and Health Informatics},
  year={2024},
  publisher={IEEE},
doi = {https://doi.org/10.1109/JBHI.2024.3471928},
url = {https://ieeexplore.ieee.org/abstract/document/10713291}
}

@article{cranmer2023interpretable,
  title={Interpretable machine learning for science with {PySR} and {SymbolicRegression.jl}},
  author={Cranmer, Miles},
  journal={arXiv preprint arXiv:2305.01582},
  year={2023},
doi = {https://doi.org/10.48550/arXiv.2305.01582},
url = {https://arxiv.org/abs/2305.01582}
}

@inproceedings{bordt2022post,
  title={Post-hoc explanations fail to achieve their purpose in adversarial contexts},
  author={Bordt, Sebastian and Finck, Mich{\`e}le and Raidl, Eric and Von Luxburg, Ulrike},
  booktitle={Proceedings of the 2022 ACM Conference on Fairness, Accountability, and Transparency},
url = {https://dl.acm.org/doi/abs/10.1145/3531146.3533153},
  pages={891--905},
  year={2022}
}

@article{camburu2020explaining,
  title={Explaining deep neural networks},
  author={Camburu, Oana-Maria},
  journal={arXiv preprint arXiv:2010.01496},
url = {https://arxiv.org/abs/2010.01496},
  year={2020}
}

@article{cheon2024kolmogorov,
  title={Kolmogorov-arnold network for satellite image classification in remote sensing},
  author={Cheon, Minjong},
  journal={arXiv preprint arXiv:2406.00600},
  year={2024},
doi = {https://doi.org/10.48550/arXiv.2406.00600},
url = {https://arxiv.org/abs/2406.00600}
}

@article{seydi2024unveiling,
  title={Unveiling the power of wavelets: A wavelet-based kolmogorov-arnold network for hyperspectral image classification},
  author={Seydi, Seyd Teymoor and Bozorgasl, Zavareh and Chen, Hao},
  journal={arXiv preprint arXiv:2406.07869},
  year={2024},
doi = {https://doi.org/10.48550/arXiv.2406.07869},
url = {https://arxiv.org/abs/2406.07869}
}

@article{yu2024kan,
  title={{KAN} or {MLP}: A fairer comparison},
  author={Yu, Runpeng and Yu, Weihao and Wang, Xinchao},
  journal={arXiv preprint arXiv:2407.16674},
  year={2024},
doi = {https://doi.org/10.48550/arXiv.2407.16674},
url = {https://arxiv.org/abs/2407.16674}
}

@article{xu2024fourierkan,
  title={Fourier{KAN}-{GCF}: Fourier Kolmogorov-Arnold Network--An Effective and Efficient Feature Transformation for Graph Collaborative Filtering},
  author={Xu, Jinfeng and Chen, Zheyu and Li, Jinze and Yang, Shuo and Wang, Wei and Hu, Xiping and Ngai, Edith C-H},
  journal={arXiv preprint arXiv:2406.01034},
  year={2024},
doi = {https://doi.org/10.48550/arXiv.2406.01034},
url = {https://arxiv.org/abs/2406.01034}
}

@article{shannon1948mathematical,
  title={A mathematical theory of communication},
  author={Shannon, Claude E},
  journal={The Bell system technical journal},
  volume={27},
  number={3},
  pages={379--423},
  year={1948},
  publisher={Nokia Bell Labs},
doi = {https://doi.org/10.1002/j.1538-7305.1948.tb01338.x},
url = {https://ieeexplore.ieee.org/abstract/document/6773024}
}

@article{faroughi2025neural,
  title={Neural Tangent Kernel Analysis to Probe Convergence in Physics-informed Neural Solvers: {PIKAN}s vs. {PINN}s},
  author={Faroughi, Salah A and Mostajeran, Farinaz},
  journal={arXiv preprint arXiv:2506.07958},
  year={2025},
doi = {https://doi.org/10.48550/arXiv.2506.07958},
url = {https://arxiv.org/abs/2506.07958}
}

@article{cuomo2022scientific,
  title={Scientific machine learning through physics--informed neural networks: Where we are and what’s next},
  author={Cuomo, Salvatore and Di Cola, Vincenzo Schiano and Giampaolo, Fabio and Rozza, Gianluigi and Raissi, Maziar and Piccialli, Francesco},
  journal={Journal of Scientific Computing},
  volume={92},
  number={3},
  pages={88},
  year={2022},
  publisher={Springer},
doi = {https://doi.org/10.1007/s10915-022-01939-z},
url = {https://link.springer.com/article/10.1007/S10915-022-01939-Z}
}

@article{MAO2020112789,
title = {Physics-informed neural networks for high-speed flows},
journal = {Computer Methods in Applied Mechanics and Engineering},
volume = {360},
pages = {112789},
year = {2020},
issn = {0045-7825},
doi = {https://doi.org/10.1016/j.cma.2019.112789},
url = {https://www.sciencedirect.com/science/article/pii/S0045782519306814},
author = {Zhiping Mao and Ameya D. Jagtap and George Em Karniadakis}
}

@article{JAGTAP2022111402,
title = {Physics-informed neural networks for inverse problems in supersonic flows},
journal = {Journal of Computational Physics},
volume = {466},
pages = {111402},
year = {2022},
issn = {0021-9991},
doi = {https://doi.org/10.1016/j.jcp.2022.111402},
url = {https://www.sciencedirect.com/science/article/pii/S0021999122004648},
author = {Ameya D. Jagtap and Zhiping Mao and Nikolaus Adams and George Em Karniadakis}
}

@article{xiapikans,
  title={Pikans: Physics-Informed Kolmogorov-Arnold Networks for Landslide Time-to-Failure Prediction},
  author={Xia, Na and Jian, Ziheng and Wu, Jinhua and Li, JingYang and Wang, Kai and Lian, Mengqi and others},
  year={2025},
journal={SSRN},
  doi = {http://dx.doi.org/10.2139/ssrn.5312086},
  url = {https://ssrn.com/abstract=5312086}
}

@article{poluektov2023construction,
  title={Construction of the Kolmogorov-Arnold representation using the Newton-Kaczmarz method},
  author={Poluektov, Michael and Polar, Andrew},
  journal={arXiv preprint arXiv:2305.08194},
  year={2023},
  doi = {https://doi.org/10.48550/arXiv.2305.08194},
  url = {https://arxiv.org/abs/2305.08194}
}

@article{cruz2025state,
  title={State-Space Kolmogorov Arnold Networks for Interpretable Nonlinear System Identification},
  author={Cruz, G and Renczes, Bal{\'a}zs and Runacres, Mark C and Decuyper, Jan},
  journal={IEEE Control Systems Letters},
  year={2025},
  publisher={IEEE},
  doi = {https://doi.org/10.1109/LCSYS.2025.3578019},
  url = {https://ieeexplore.ieee.org/abstract/document/11029059}
}

@article{ismailov2014approximation,
  title={On the approximation by neural networks with bounded number of neurons in hidden layers},
  author={Ismailov, Vugar E},
  journal={Journal of Mathematical Analysis and Applications},
  volume={417},
  number={2},
  pages={963--969},
  year={2014},
  publisher={Elsevier},
  doi = {https://doi.org/10.1016/j.jmaa.2014.03.092},
url={https://www.sciencedirect.com/science/article/pii/S0022247X14003412}
}

@article{ismailov2023three,
  title={A three layer neural network can represent any multivariate function},
  author={Ismailov, Vugar E},
  journal={Journal of Mathematical Analysis and Applications},
  volume={523},
  number={1},
  pages={127096},
  year={2023},
  publisher={Elsevier},
  doi = {https://doi.org/10.1016/j.jmaa.2023.127096},
url={https://www.sciencedirect.com/science/article/abs/pii/S0022247X23000999}
}

@article{zeng2024kan,
  title={Kan versus mlp on irregular or noisy functions},
  author={Zeng, Chen and Wang, Jiahui and Shen, Haoran and Wang, Qiao},
  journal={arXiv preprint arXiv:2408.07906},
  year={2024},
doi = {https://doi.org/10.48550/arXiv.2408.07906},
url={https://arxiv.org/abs/2408.07906}
}

@article{zhou2025askan,
  title={as{KAN}: Active Subspace embedded Kolmogorov-Arnold Network},
  author={Zhou, Zhiteng and Xu, Zhaoyue and Liu, Yi and Wang, Shizhao},
  journal={arXiv preprint arXiv:2504.04669},
  year={2025},
doi = {https://doi.org/10.48550/arXiv.2504.04669},
url={https://arxiv.org/abs/2504.04669}
}

@article{Quade2016,
  title={Prediction of dynamical systems by symbolic regression},
  author={Quade, Markus and Abel, Markus and Shafi, Kamran and Niven, Robert K and Noack, Bernd R},
  journal={Physical Review E},
  volume={94},
  number={1},
  pages={012214},
  year={2016},
doi = {https://doi.org/10.1103/PhysRevE.94.012214},
  publisher={APS}
}

@article{Gondhalekar2009,
  title={Parameters identification for nonlinear dynamic systems via genetic algorithm optimization},
  author={Gondhalekar, AC and Petrov, EP and Imregun, M},
  journal={Journal of Computational and Nonlinear Dynamics},
 volume={4},
  number={4},
doi = {https://doi.org/10.1115/1.3187213},
  year={2009}
}

@article{Ljung2010,
  title={Perspectives on system identification},
  author={Ljung, Lennart},
  journal={Annual Reviews in Control},
  volume={34},
  number={1},
  pages={1--12},
doi = {https://doi.org/10.1016/j.arcontrol.2009.12.001},
  year={2010},
  publisher={Elsevier}
}

@article{Brunton2016,
  title={Discovering governing equations from data by sparse identification of nonlinear dynamical systems},
  author={Brunton, Steven L and Proctor, Joshua L and Kutz, J Nathan},
  journal={Proceedings of the national academy of sciences},
  volume={113},
  number={15},
  pages={3932--3937},
  year={2016},
doi = {https://doi.org/10.1073/pnas.1517384113},
  publisher={National Academy of Sciences}
}

@article{ADAMSINDy,
  title={Adam-sindy: An efficient optimization framework for parameterized nonlinear dynamical system identification},
  author={Viknesh, Siva and Tatari, Younes and Christenson, Chase and Arzani, Amirhossein},
 journal={Physical Review Research},
  volume={8},
  number={1},
  pages={013040},
  year={2026}
}

@article{Samek2021,
  title={Explaining deep neural networks and beyond: A review of methods and applications},
  author={Samek, Wojciech and Montavon, Gr{\'e}goire and Lapuschkin, Sebastian and Anders, Christopher J and M{\"u}ller, Klaus-Robert},
  journal={Proceedings of the IEEE},
  volume={109},
  number={3},
  pages={247--278},
doi={https://doi.org/10.1109/JPROC.2021.3060483},
  year={2021},
  publisher={IEEE}
}

@article{faroughi2026KanRev,
  title={Kolmogorov-Arnold networks for data-driven, physics-informed, and deep-operator learning: a review, synthesis, and new analysis},
  author={Faroughi, Salah A and Mostajeran, Farinaz and Mashhadzadeh, Amin Hamed and Faroughi, Shirko},
  journal={Neural networks},
  volume={200},
  number={},
  pages={108791},
  year={2026},
doi={https://doi.org/10.1016/j.neunet.2026.108791},
  publisher={Elsevier}
}

@article{faroughi2024SciMLRev,
  title={Physics-guided, physics-informed, and physics-encoded neural networks and operators in scientific computing: Fluid and solid mechanics},
  author={Faroughi, Salah A and Pawar, Nikhil M and Fernandes, Celio and Raissi, Maziar and Das, Subasish and Kalantari, Nima K and Kourosh Mahjour, Seyed},
  journal={Journal of Computing and Information Science in Engineering},
  volume={24},
  number={4},
  pages={040802},
  year={2024},
doi={https://doi.org/10.1115/1.4064449},
  publisher={American Society of Mechanical Engineers}
}

@article{fukami2021sparse,
  title={Sparse identification of nonlinear dynamics with low-dimensionalized flow representations},
  author={Fukami, K. and Murata, T. and Zhang, K. and Fukagata, K.},
  journal={Journal of Fluid Mechanics},
  volume={926},
  pages={A10},
  year={2021},
  publisher={Cambridge University Press}
}

@article{yang2020rapid,
  title={Rapid data-driven model reduction of nonlinear dynamical systems including chemical reaction networks using l1-regularization},
  author={Yang, Q. and Sing-Long, C. A. and Reed, E. J.},
  journal={Chaos: An Interdisciplinary Journal of Nonlinear Science},
  volume={30},
  number={5},
  year={2020},
  publisher={AIP Publishing}
}

@article{hoffmann2019reactive,
  title={Reactive {SINDy}: Discovering governing reactions from concentration data},
  author={Hoffmann, M. and Fr{\"o}hner, C. and No{\'e}, F.},
  journal={The Journal of Chemical Physics},
  volume={150},
  number={2},
  year={2019},
  publisher={AIP Publishing}
}

@article{kaptanoglu2023benchmarking,
  title={Benchmarking sparse system identification with low-dimensional chaos},
  author={Kaptanoglu, A. A. and Zhang, L. and Nicolaou, Z. G. and Fasel, U. and Brunton, S. L.},
  journal={Nonlinear Dynamics},
  volume={111},
  number={14},
  pages={13143--13164},
  year={2023},
  publisher={Springer}
}

@article{fasel2022ensemble,
  title={{Ensemble-SINDy}: Robust sparse model discovery in the low-data, high-noise limit, with active learning and control},
  author={Fasel, U. and Kutz, J. N. and Brunton, B. W. and Brunton, S. L.},
  journal={Proceedings of the Royal Society A: Mathematical, Physical and Engineering Sciences},
  volume={478},
  number={2260},
  year={2022},
  publisher={The Royal Society}
}

@article{boninsegna2018sparse,
  title={Sparse learning of stochastic dynamical equations},
  author={Boninsegna, L. and N{\"u}ske, F. and Clementi, C.},
  journal={The Journal of Chemical Physics},
  volume={148},
  number={24},
  year={2018},
  publisher={AIP Publishing}
}

@article{champneys2025bindy,
  title={{BINDy}: Bayesian identification of nonlinear dynamics with reversible-jump {Markov-chain Monte Carlo}},
  author={Champneys, M. D. and Rogers, T. J.},
  journal={Proceedings of the Royal Society A: Mathematical, Physical and Engineering Sciences},
  volume={481},
  number={2319},
  year={2025},
  publisher={The Royal Society}
}

@article{arzani2025interpreting,
  title={Interpreting and generalizing deep learning in physics-based problems with functional linear models},
  author={Arzani, A. and Yuan, L. and Newell, P. and Wang, B.},
  journal={Engineering with Computers},
  volume={41},
  number={1},
  pages={135--157},
  year={2025},
  publisher={Springer}
}

@article{csala2026decomposed,
  title={Decomposed sparse modal optimization: Interpretable reduced-order modeling of unsteady flows},
  author={Csala, H. and Arzani, A.},
  journal={International Journal of Heat and Fluid Flow},
  volume={117},
  pages={110124},
  year={2026},
  publisher={Elsevier}
}

@article{brunton2025machine,
  title={Machine learning for sparse nonlinear modeling and control},
  author={Brunton, S. L. and Zolman, N. and Kutz, J. N. and Fasel, U.},
  journal={Annual Review of Control, Robotics, and Autonomous Systems},
  volume={8},
  number={1},
  pages={127--152},
  year={2025},
  publisher={Annual Reviews}
}

@article{yonezawa2026sparse,
  title={Sparse identification of nonlinear dynamics with library optimization mechanism: Recursive long-term prediction perspective},
  author={Yonezawa, A. and Yonezawa, H. and Yahagi, S. and Kajiwara, I.  and Kijimoto, S. and Taniuchi, H. and Murakami, K.},
  journal={IEEE Transactions on Cybernetics},
  year={2026},
  publisher={IEEE}
}

\end{document}